%% file: colm2026_conference.tex
\documentclass{article} 
\usepackage[preprint]{colm2026_conference}

\usepackage{microtype}
\usepackage{hyperref}
\usepackage{url}
\usepackage{booktabs}
\usepackage{siunitx}
\usepackage{multirow}
\usepackage{graphicx}
\usepackage{tabularx}
\usepackage{xspace}
\usepackage{enumitem}
\usepackage{amssymb}
\usepackage{xcolor}
\usepackage{pifont}
\usepackage{caption}
\usepackage{float}
\newcommand{\cmark}{{\color{green!70!black}\ding{51}}}
\newcommand{\xmark}{{\color{red!80!black}\ding{55}}}

\definecolor{darkblue}{rgb}{0, 0, 0.5}
\hypersetup{colorlinks=true, citecolor=darkblue, linkcolor=darkblue, urlcolor=darkblue}

\title{iOSWorld: A Benchmark for Personally Intelligent Phone Agents}

\author{Lawrence Keunho Jang,\;
Mareks Woodside\thanks{Equal contribution.},\;
Geronimo Carom\footnotemark[1],\\
\textbf{Andrew Keunwoo Jang}\footnotemark[1], \;
\textbf{Jing Yu Koh},\;
\textbf{Ruslan Salakhutdinov} \\
Carnegie Mellon University \\
\texttt{\{ljang, rsalakhu\}@cs.cmu.edu}
}

\def\iosworld{iOSWorld\xspace}

\begin{document}

\maketitle

\begin{abstract}
A useful phone agent needs to be personally intelligent. It should reason over a user's identity, history, and preferences as they exist on the device, not just follow isolated instructions in an impersonal sandbox. Existing mobile agent benchmarks lack this kind of personalization. We introduce iOSWorld, the first interactive native iOS simulator benchmark built around a persistent user identity spanning 26 newly built iOS apps. These apps contain connected data such as transactions, messages, travel records, social relationships, and financial activity. iOSWorld includes 133 tasks across three increasingly difficult categories. Single-app tasks (27) test one app, multi-app tasks (60) span 2 to 8 apps, and memory and personalization tasks (46) require agents to infer patterns from personal data. We evaluate frontier and open-source computer-use models in both vision-only and privileged vision+XML settings. The best configuration reaches 52\% overall but only 37\% on multi-app tasks. Privileged vision+XML access improves frontier models by up to 26 percentage points, while smaller models do not benefit from added accessibility-tree input. We release iOSWorld as an open-source benchmark with all apps, seeded data, tasks, rubrics, and evaluation code at \url{https://iosworld.io}.
\end{abstract}

\section{Introduction}

A person's phone is not a blank slate. Transactions, messages, social connections, and financial records accumulate across many applications, forming a record that any useful assistant has to understand and navigate. We call the corresponding agent capability \emph{personally intelligent}: reasoning over a user's identity, history, and preferences as they exist on the device, rather than executing sandboxed, isolated tasks. Current phone-agent benchmarks ignore this dimension. Tasks are issued against app states with no persistent user data, no cross-app continuity, and no notion of a real user. An agent that taps the right button on a settings screen but cannot find its owner's most common commute route has not shown useful capability.

Existing benchmarks evaluate digital agents on Android~\citep{rawles2025androidworld, kong2025mobileworldbenchmarkingautonomousmobile}, web~\citep{zhou2023webarena,koh2024visualwebarena}, and desktop~\citep{xie2024osworld, yang2025macosworld, bonatti2024windows}. iOS serves over 2.5 billion active devices\footnote{Apple installed-base figure, 2026: \url{https://finance.yahoo.com/news/apple-installed-tops-2-5-170414353.html}} and 58--60\% of U.S. mobile OS usage\footnote{StatCounter Global Stats (accessed March 2026): \url{https://gs.statcounter.com/os-market-share/mobile/united-states}}, yet interactive phone-agent benchmarks target Android, not native iOS. None populate apps with persistent user identity. We exclude tasks centered on web browsing, since web agents are already well-served by existing work~\citep{zhou2023webarena,koh2024visualwebarena,he2024webvoyager} and a phone agent with browser access can be evaluated there directly. Our focus is on native iOS apps and the personal data they hold.

\iosworld is the first dynamic native iOS simulator benchmark built around a user's personal identity. We built 26 native iOS applications and populated them with connected data for a single persona, Jordan Avery. The same contacts appear across messaging, payment, and email. A food order on one app produces a bank charge and a receipt email in others. An upcoming flight matches a hotel booking and confirmation emails across separate apps. We release \textbf{133 tasks} in three categories. \textbf{Single-app} tasks (27) test basic interaction within one app. \textbf{Multi-app} tasks (60) carry information across 2 to 8 applications. \textbf{Memory and personalization} tasks (46) require agents to discover implicit patterns from in-app data without being told where to look. The release includes a schema for adding new tasks and seeding personalized data. Our contributions:

\begin{itemize}[nosep,leftmargin=*]
    \item The first interactive native iOS simulator benchmark with one user identity spanning 26 purpose-built applications containing connected personal data.
    \item 133 tasks across three categories, evaluated with an LLM-as-a-judge pipeline validated against human annotators ($\kappa$=0.77).
    \item A comparison of five frontier models and one open-source baseline (Qwen3.5 35B-A3B) under vision-only and privileged vision+XML settings. The best overall configuration achieves 52\% overall (82\% single-app, 54\% memory, 37\% multi-app). Privileged vision+XML access improves the stronger frontier models by up to 26 percentage points, while smaller models do not show the same gain.
    \item We open-source all apps, seed data, tasks, rubrics, and evaluation code, plus an AWS-runner (EC2-managed Mac instances) so non-Mac researchers can run the benchmark. Code at \url{https://github.com/ljang0/iOSWorld} and site at \url{https://iosworld.io}.
\end{itemize}

\begin{figure*}[t]
\vspace{-0.4in}
\centering
\includegraphics[width=\textwidth]{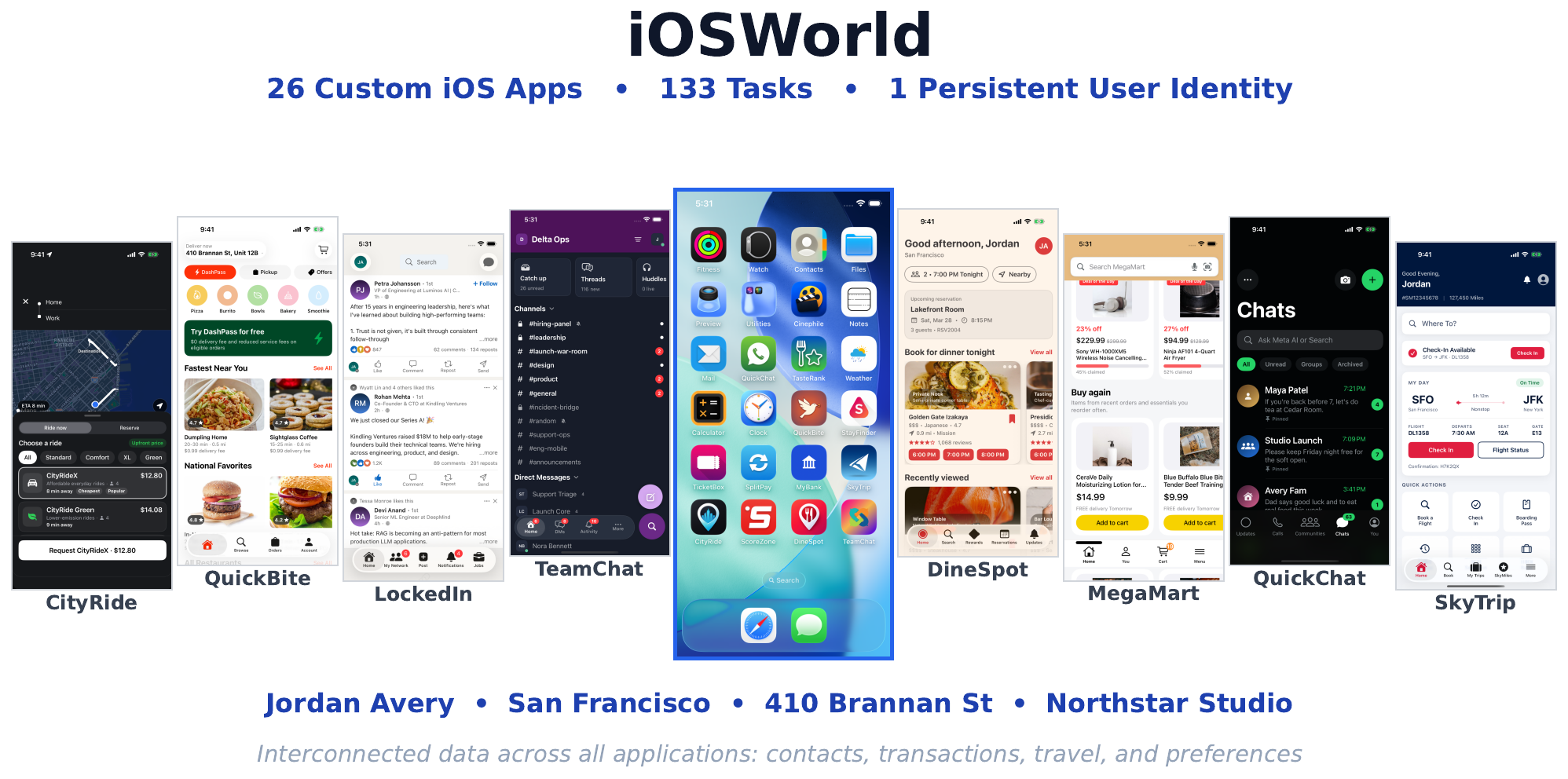}
\caption{\small Overview of \iosworld. 26 purpose-built iOS applications share a single user identity (Jordan Avery) and connected data across apps. The benchmark includes 133 tasks across single-app, multi-app, and memory/personalization categories.}
\label{fig:overview}
\vspace{-0.2in}
\end{figure*}

\section{Related Work}

\subsection{GUI Agent Benchmarks}

GUI Agents have largely been centered on the web and desktop. Foundational benchmarks such as MiniWoB, MiniWoB++, WebShop, WebArena, and VisualWebArena~\citep{shi2017world,liu2018reinforcement,yao2022webshop, zhou2023webarena, koh2024visualwebarena} emulated tasks on the web. Mind2Web~\citep{deng2023mind2web} and WebVoyager~\citep{he2024webvoyager} extended to real websites, and \citet{xue2025illusion} studied how evaluations transfer to live conditions. At the OS and Desktop level, OSWorld~\citep{xie2024osworld} covers Linux, Windows Agent Arena~\citep{bonatti2024windows} targets Windows, MacOSWorld~\citep{yang2025macosworld} covers macOS, and WorkArena~\citep{drouin2024workarena} benchmarks enterprise knowledge work. GAIA~\citep{mialon2023gaia}, TheAgentCompany~\citep{xu2025theagentcompany}, and $\tau$-bench~\citep{yao2024taubench} probe multi-step and multi-tool reasoning. All of these benchmarks present agents on desktops with predominantly impersonal or single user environments and explicit instructions.
\vspace{-0.2in}
\subsection{Mobile Device Agents}

Existing interactive mobile-agent benchmarks target Android. AndroidEnv~\citep{toyama2021androidenv} provides an RL interface for phone agents, Android-in-the-Wild~\citep{rawles2024aitw} provides evaluation using human demonstrations on Android, and AndroidWorld~\citep{rawles2025androidworld} offers dynamic tasks with programmatic verification on a live Android simulator. MobileWorld~\citep{kong2025mobileworldbenchmarkingautonomousmobile} extends AndroidWorld with long-horizon and tool use. Additional benchmarks have expanded coverage across different digital domains, such as AndroidLab~\citep{xu2024androidlab}, SPA-Bench~\citep{chen2025spabench}, B-MoCA~\citep{lee2025bmoca}, and GUI Odyssey~\citep{lu2025guiodyssey}. On the modeling side, CogAgent~\citep{hong2023cogagent}, AppAgent~\citep{yang2023appagent}, Mobile-Agent~\citep{wang2024mobileagent}, UI-TARS~\citep{qin2025uitars}, and AutoDroid~\citep{wen2024autodroid} explore architectures ranging from fine-tuned VLMs to RL-trained agents~\citep{bai2024digirl,bai2025digiq} for mobile agents. Earlier work on inducing mobile skills from user demonstrations~\citep{shen2019skillbot} pre-dates the LLM-agent era. Ferret-UI~\citep{you2024ferretui} develops a multimodal model for understanding mobile UI on both Android and iOS.

There remains a gap for dynamic iOS evaluations. iOS differs from Android in its UI framework, navigation patterns, and accessibility infrastructure. No mobile benchmark on any platform seeds applications with a user identity or evaluates reasoning over extensive personal data distributed across apps.



\begin{figure}[H]
\vspace{-0.05in}
\centering
\includegraphics[width=\textwidth]{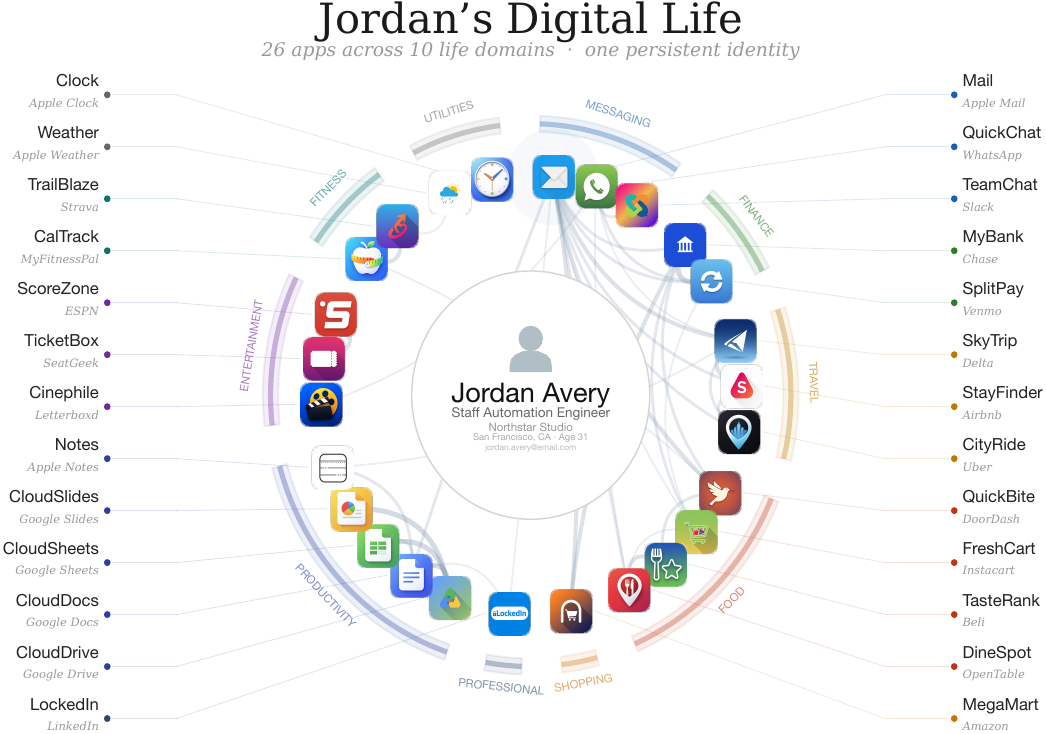}
\vspace{-0.1in}
\caption{\small Jordan Avery's digital life. 26 iOS apps across 10 domains sharing one identity. We display app names in \textbf{bold} and real-life analogues in \textit{italics}. Edge thickness represents the number of shared data points; Mail is the primary hub. See Table~\ref{tab:apps} for details on the applications.}
\label{fig:jordans-life}
\vspace{-0.15in}
\end{figure}

\section{\iosworld}

\subsection{Environment}

We model the \iosworld environment as a partially observable Markov decision process (POMDP)~\citep{kaelbling1998pomdp}: $\mathcal{E} = (\mathcal{S}, \mathcal{A}, \Omega, T)$, where $\mathcal{S}$ is the set of simulator states, $\mathcal{A}$ is the action space, $\Omega$ is the observation space, and $T: \mathcal{S} \times \mathcal{A} \rightarrow \mathcal{S}$ defines deterministic transitions. At each step $t$, the agent receives a partial observation $o_t \in \Omega$ of state $s_t$ and produces an action $a_t \in \mathcal{A}$, which transitions the simulator to $s_{t+1}$.

\paragraph{Observation space.} We evaluate agents under two observation modalities. Unlike Android, where tools like UIAutomator expose accessibility data, iOS is closed-source. The richest structured UI data available to third-party tools comes through Apple's XCUITest framework, which requires a Mac running Xcode. A deployed agent without privileged information would have access only to what is visible on the screen. We evaluate both settings to separate visual grounding, reasoning, and privileged access.

In the \textbf{vision-only} setting, the agent receives a screenshot at each step. The raw simulation captures are 1206$\times$2622; we resize to 706$\times$1536. This 1536-pixel cap on the longest edge stays within Anthropic's 1568-pixel API limit for Claude Computer Use, and we apply it uniformly across all providers for a fair comparison. The agent must visually identify UI elements, estimate their coordinates, and infer the application state from pixels alone. We do not evaluate in text-only mode since all frontier computer-use models require image input.

In the \textbf{vision+XML} setting, the agent additionally receives a cleaned accessibility tree in XML extracted via XCUITest. For each interactive element, the tree reports the element type (e.g., \texttt{Button}, \texttt{TextField}, \texttt{Cell}), display name, label, current value, center coordinates in a normalized 0--1000 space, and an accessibility identifier when available. The tree is filtered to interactive and visible elements, capped at 200 elements and 15 levels of depth.

\paragraph{Action space.} The available actions differ by modality and provider adapter (Table~\ref{tab:action-space}). In vision-only mode, agents are limited to six actions and must estimate all tap coordinates from screenshots. In vision+XML mode, action adapters expose additional tools. The most useful are \texttt{tap}, which targets elements by accessibility identifier, and \texttt{launch\_app}, which opens apps through their bundle identifier instead of visual home-screen navigation. Table~\ref{tab:action-translation} gives the exact mapping of each provider adapter.

\begin{table}[H]
\centering
\small
\begin{tabular}{@{}lll@{}}
\toprule
\textbf{Action} & \textbf{Parameters} & \textbf{Mode} \\
\midrule
\texttt{tap\_xy} & $x, y \in [0, 1000]$ & Vision, Vision+XML\\
\texttt{type} & text string & Vision, Vision+XML \\
\texttt{swipe} & direction, optional origin & Vision, Vision+XML \\
\texttt{home} & --- & Vision, Vision+XML \\
\texttt{wait} & duration (seconds) & Vision, Vision+XML \\
\texttt{stop} & answer string & Vision, Vision+XML \\
\midrule
\texttt{tap} & accessibility identifier & Vision+XML only \\
\texttt{launch\_app} & bundle identifier & Vision+XML only \\
\texttt{terminate\_app} & bundle identifier & Vision+XML only \\
\texttt{open\_url} & URL string & Vision+XML only \\
\bottomrule
\end{tabular}
\caption{\small Action space. The top block is available in both modalities, while the bottom block requires the accessibility tree and is exposed when supported by the model provider-specific action adapter (Table~\ref{tab:action-translation}). The \texttt{tap} action targets elements by identifier, enabling pixel-perfect interaction without coordinate estimation.}
\label{tab:action-space}
\end{table}

\paragraph{Translating computer-use agents (CUAs) to iOS.} Frontier computer-use models each define their own desktop action space. We adapt them to iOS with one translation layer. Click becomes \texttt{tap\_xy}, scroll becomes \texttt{swipe} with inverted direction, and coordinates are normalized to 0--1000. All models also receive iOS-specific system prompts for touchscreen-only interaction. For the open-source Qwen3.5 baseline, we follow the official Qwen3-VL mobile-agent cookbook. This exposes the cookbook's \texttt{mobile\_use} tool (\texttt{click}, \texttt{long\_press}, \texttt{swipe}, \texttt{type}, \texttt{system\_button}, \texttt{wait}, \texttt{terminate}) on a 999$\times$999 grid that we rescale to our 0--1000 schema. The full action mapping is in Table~\ref{tab:action-translation}.

\begin{figure*}[t]
\centering
\includegraphics[width=\textwidth]{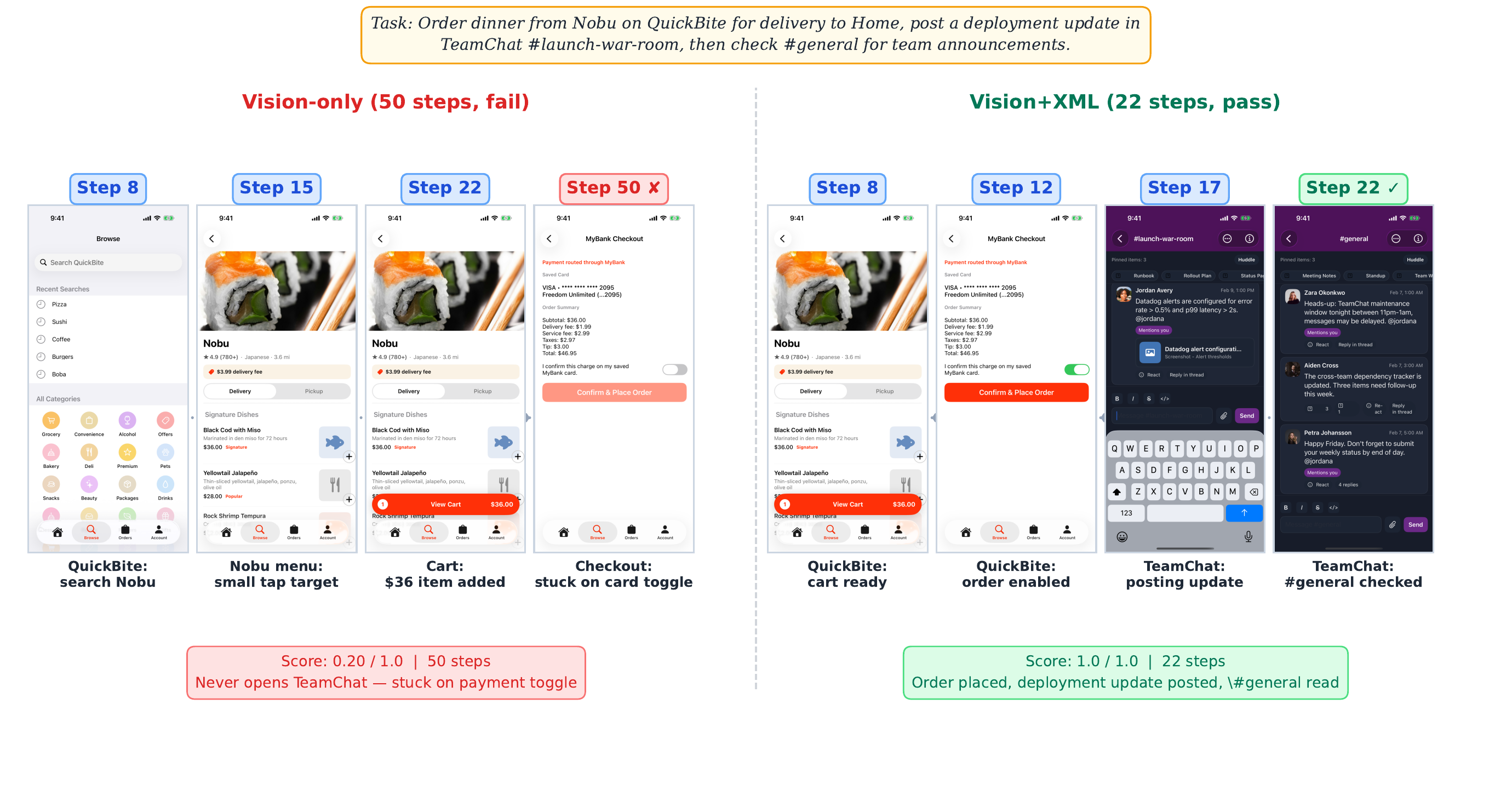}
\caption{\small
We visualize a multi-app QuickBite $\rightarrow$ TeamChat task across two modalities with the same model (Opus~4.6). \textbf{Vision-only} finds Nobu, adds an item, and reaches checkout, but spends the remaining budget trying to toggle payment confirmation and never opens TeamChat (50 steps, score 0.20). \textbf{Vision+XML} places the order, posts a deployment update in \#launch-war-room, and checks \#general announcements in 22 steps (score 1.0).}
\vspace{-0.15in}
\label{fig:xml-advantage}
\end{figure*}

\paragraph{Infrastructure.} The agent loop runs on macOS using Appium with the XCUITest driver, controlling an Xcode-managed iPhone simulator. Each task starts from a deterministic home state with all 26 apps pre-installed and seeded. The loop captures observations, sends them to the model, executes the returned actions, and repeats until the agent issues \texttt{stop} or reaches the step limit (Fig.~\ref{fig:xml-advantage} shows two same-task runs under two different modalities). Each cloned simulator instance uses 2--4 GB of RAM. Accounting for OS overhead (${\sim}$8 GB), authors found that a 36 GB Mac Studio (M4 Max) safely supports 8 parallel workers and a 24 GB MacBook Pro (M4) supports 4. Workers are managed via \texttt{xcrun simctl clone}, with each clone receiving dedicated Appium and WebDriverAgent ports.

\paragraph{Evaluation.} Each task is scored with an LLM-as-a-Judge \citep{zheng2023judgingllmasajudgemtbenchchatbot} framework using GPT-5.4-Mini. The judge reviews the full trajectory, including screenshots, actions, and the final answer, then returns a binary pass/fail judgment. Human validation on 128 Opus~4.6 trajectories confirms substantial agreement ($\kappa$=0.77 at task level, 89\% accuracy; see \S\ref{sec:analysis}). We also tested a per-step variant that evaluates screenshots independently, but it was more lenient without improving discrimination. We use the trajectory-level judge throughout and report binary pass rate as the primary metric. Details on per-step evaluation and rubric scoring are in Appendix~\ref{app:rubric-details}.

\subsection{App Ecosystem and User Identity}

All 26 applications share a single user identity: \textbf{Jordan Avery}, a San Francisco-based professional living at 410 Brannan Street who works at Northstar Studio and trains for a half marathon (Fig.~\ref{fig:jordans-life}). Jordan's contacts, Maya Patel, Leo Chen, Kai Santos, appear as QuickChat correspondents, SplitPay payees, Mail senders, LockedIn connections, and TeamChat colleagues. A Chipotle order in QuickBite produces a charge in MyBank and a receipt in Mail. An upcoming SFO$\rightarrow$JFK flight in SkyTrip aligns with a StayFinder booking and a Notes reminder. These cross-references make multi-app and memory tasks require evidence from more than one application.

Apps were developed or adapted in SwiftUI using Claude Code as a coding assistant, then manually verified by human developers for correct navigation, data rendering, and seed data consistency. The applications implement tab-based navigation, searchable lists, detail views, and editing flows that follow their real-world counterparts. Two apps build on open-source foundations: Notes is based on snowNotes\footnote{\url{https://github.com/probablyhades/snowNotes}} and Cinephile draws from MovieSwiftUI\footnote{\url{https://github.com/Dimillian/MovieSwiftUI}}. User data is encoded in Swift seed fixtures and JSON snapshots loaded at build time. The 26 apps span finance, messaging, travel, food, shopping, productivity, entertainment, fitness, utilities, and professional networking. Full details are in Table~\ref{tab:apps}.

\subsection{Task Design}

\iosworld includes 133 tasks in three increasingly difficult categories (Table~\ref{tab:task-examples}). \textbf{Single-app tasks} (27) test basic navigation and interaction within one app, such as logging a meal in CalTrack or finding an upcoming flight in SkyTrip. \textbf{Multi-app tasks} (60) span two to eight applications and require transferring information between them. For example, one task asks the agent to check a Chipotle order on QuickBite, find the matching charge in MyBank, locate the receipt email in Mail, and note any price differences in Notes. \textbf{Memory and personalization tasks} (46) require discovering latent patterns that are never stated. The agent is asked questions like ``What is my most common commute route?'' or ``Find my most frequently ordered restaurant and place a reorder.'' Correct answers require exploration, pattern finding, and synthesis across apps.

\begin{table*}[t]
\vspace{-0.3in}
\centering
\small
\resizebox{\textwidth}{!}{%
\begin{tabular}{@{}llp{9.5cm}l@{}}
\toprule
\textbf{Category} & \textbf{Task ID} & \textbf{Task Instruction} & \textbf{Apps} \\
\midrule
\multirow[c]{2}{*}{Single-App}
& dinespot-001 & Search for restaurants in San Francisco with ``Outdoor Seating'' on DineSpot and make a reservation at Harborline Seafood for 2 tonight at 7 PM. & DineSpot \\
\cmidrule(lr){2-4}
& quickchat-003 & Search my QuickChat for ``Brooklyn Half'' and find which conversation mentioned it. Reply to that thread confirming I'm registered. & QuickChat \\
\midrule
\multirow[c]{2}{*}{Multi-App}
& multi-009 & Check my most recent Chipotle order on QuickBite. Then check my MyBank credit card for the corresponding charge. Find the receipt email in Mail and note any price differences in Notes. & QuickBite, MyBank, Mail, Notes \\
\cmidrule(lr){2-4}
& multi-011 & Check my StayFinder trip for Catalina Island (Apr 18--21). Look up the weather for those dates. Check my TasteRank ``Want to Try'' list for nearby restaurants and compile everything in Notes. & StayFinder, Weather, TasteRank, Notes \\
\midrule
\multirow[c]{2}{*}{Memory}
& mem-002 & Look at my CityRide app and figure out my most common route based on my saved locations. Then request a ride along that route. & CityRide \\
\cmidrule(lr){2-4}
& mem-005 & Review my TrailBlaze activities to figure out my regular running schedule and favorite routes. Check the Weather for conditions during my typical run time and message my running group. & TrailBlaze, Weather, QuickChat \\
\bottomrule
\end{tabular}%
}
\caption{\small Example tasks from each category.}
\vspace{-0.15in}
\label{tab:task-examples}
\end{table*}

\paragraph{Task creation.} We generated tasks using Claude Code \citep{anthropic2026ClaudeCode} with full access to each app's source code and seed data. The coding agent examined seeded JSON files, view controllers, and navigation flows, then produced tasks grounded in the actual app state. Each task is written in first-person voice and accompanied by rubric criteria that decompose the objective into verifiable steps (Appendix~\ref{app:rubric-details}). Human annotators reviewed and refined every task that required changes.

\paragraph{Quality assurance.} Grounding tasks in seed data required careful verification. Human annotators manually executed every task end-to-end on the iOS simulator and verified feasibility. Forty-four of the 175 candidate tasks required corrections, including nonexistent flight routes, mismatched food names, and rubric criteria that referenced unreachable app states. All 26 apps were independently tested for UI elements, seed data, and navigation flows.

The initial pool contained 175 tasks. We trimmed the single-app set for broad app coverage with minimal duplication and kept all multi-app and memory tasks, leaving a final set of 133 tasks. Memory tasks involve 4.4 apps per task on average, since answering them requires exploring several data sources. QuickChat appears in 44 of 133 tasks and Notes in 41, with CloudDocs at 35 and Mail at 29, making them the most frequently referenced apps.

\section{Experiments}

\subsection{Setup}

We evaluate five frontier computer-use models: Claude Opus 4.6 and Claude Sonnet 4.6 \citep{anthropic2026opus46}, GPT-5.4 and GPT-5.4 Mini \citep{openai2026gpt54}, and Gemini 3 Flash \citep{google2026gemini}. Each provider offers a dedicated computer-use API with native screenshot understanding and action generation. We also include Qwen3.5 35B-A3B \citep{qwen2026qwen35}, an open-weights mixture-of-experts model with 35B total and 3B active parameters, served via vLLM and prompted with the official Qwen3-VL mobile-agent cookbook. We test each model under both Vision-only and Vision+XML, yielding twelve configurations. All runs use a 50-step limit and screenshots capped at 1536 pixels on the longest edge. Within each modality, all models receive equivalent system prompts adapted to their action vocabularies. We use GPT-5.4~Mini as the trajectory judge. Human agreement analysis on 128 Opus~4.6 trajectories confirms substantial agreement ($\kappa$=0.77 task-level; Appendix~\ref{app:human-agreement}). 

\subsection{Results}

\begin{table*}[t]
\vspace{-0.1in}
\centering
\small
\resizebox{\textwidth}{!}{%
\begin{tabular}{@{}lc
                r@{\hskip 6pt}S[table-format=2.1]
                r@{\hskip 6pt}S[table-format=2.1]
                r@{\hskip 6pt}S[table-format=2.1]
                r@{\hskip 6pt}S[table-format=2.1]@{}}
\toprule
& & \multicolumn{2}{c}{\textbf{Single (27)}} & \multicolumn{2}{c}{\textbf{Multi (60)}} & \multicolumn{2}{c}{\textbf{Memory (46)}} & \multicolumn{2}{c}{\textbf{Overall (133)}} \\
\cmidrule(lr){3-4} \cmidrule(lr){5-6} \cmidrule(lr){7-8} \cmidrule(lr){9-10}
\textbf{Model} & \textbf{+XML} & {Pass} & {Steps} & {Pass} & {Steps} & {Pass} & {Steps} & {Pass} & {Steps} \\
\midrule
\multirow{2}{*}{Opus 4.6}
  & \xmark       & 70.4\% & 23.1 & 20.0\% & 45.5 & 8.7\% & 49.4 & 26.3\% & 42.3 \\
  & \cmark       & 81.5\% & 16.4 & \textbf{36.7}\% & 38.2 & \textbf{54.3}\% & 39.3 & \textbf{51.9}\% & 34.1 \\
\midrule
\multirow{2}{*}{Sonnet 4.6}
  & \xmark       & 77.8\% & 26.3 & 18.3\% & 46.9 & 13.0\% & 49.8 & 28.6\% & 43.7 \\
  & \cmark       & \textbf{92.6}\% & 15.4 & 35.0\% & 42.0 & 34.8\% & 44.8 & 46.6\% & 37.5 \\
\midrule
\multirow{2}{*}{GPT-5.4}
  & \xmark       & 63.0\% & 32.5 & 11.7\% & 48.5 & 6.5\% & 48.3 & 20.3\% & 45.2 \\
  & \cmark       & 81.5\% & 12.1 & 26.7\% & 37.1 & 32.6\% & 37.2 & 39.8\% & 32.1 \\
\midrule
\multirow{2}{*}{GPT-5.4 Mini}
  & \xmark       & 70.4\% & 24.0 & 18.3\% & 46.3 & 10.9\% & 48.9 & 26.3\% & 42.7 \\
  & \cmark       & 66.7\% & 14.9 & 1.7\% & 43.8 & 4.3\% & 41.5 & 15.8\% & 37.2 \\
\midrule
\multirow{2}{*}{Gemini 3 Flash}
  & \xmark       & 70.4\% & 11.0 & 13.3\% & 23.9 & 21.7\% & 23.5 & 27.8\% & 21.2 \\
  & \cmark       & 70.4\% & 11.5 & 18.3\% & 38.0 & 17.4\% & 33.3 & 28.6\% & 31.0 \\
\midrule
\multirow{2}{*}{Qwen3.5 35B-A3B}
  & \xmark       & 40.7\% & 23.0 & 6.7\% & 36.2 & 4.3\% & 33.4 & 12.8\% & 32.6 \\
  & \cmark       & 48.1\% & 31.4 & 0.0\% & 43.9 & 2.2\% & 37.9 & 10.5\% & 39.3 \\
\bottomrule
\end{tabular}%
}

\caption{\small Pass rates (\%) and average steps by task category. Rows with \xmark{} are vision-only. \cmark{} denotes vision+XML. Rubric-based scoring in Appendix~\ref{app:rubric-details}.}
\label{tab:main-results}
\end{table*}

\begin{figure*}[t]
\centering
\includegraphics[width=\textwidth]{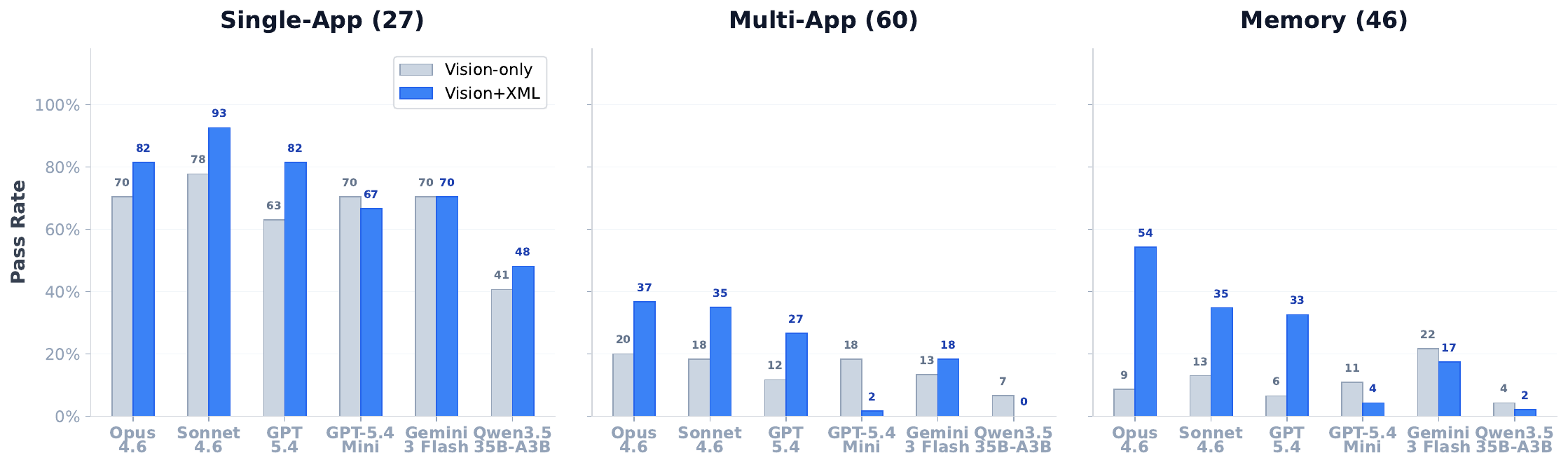}
\caption{\small Pass rates by task category and observation modality across all six models. Vision+XML (blue) outperforms vision-only (gray) for the stronger frontier models. GPT-5.4 Mini and the open-source Qwen3.5 baseline do not show the same benefit from the additional modality.}
\label{fig:results-bars}
\end{figure*}

\begin{figure*}[t]
\vspace{-0.1in}
\centering
\includegraphics[width=0.82\textwidth,keepaspectratio]{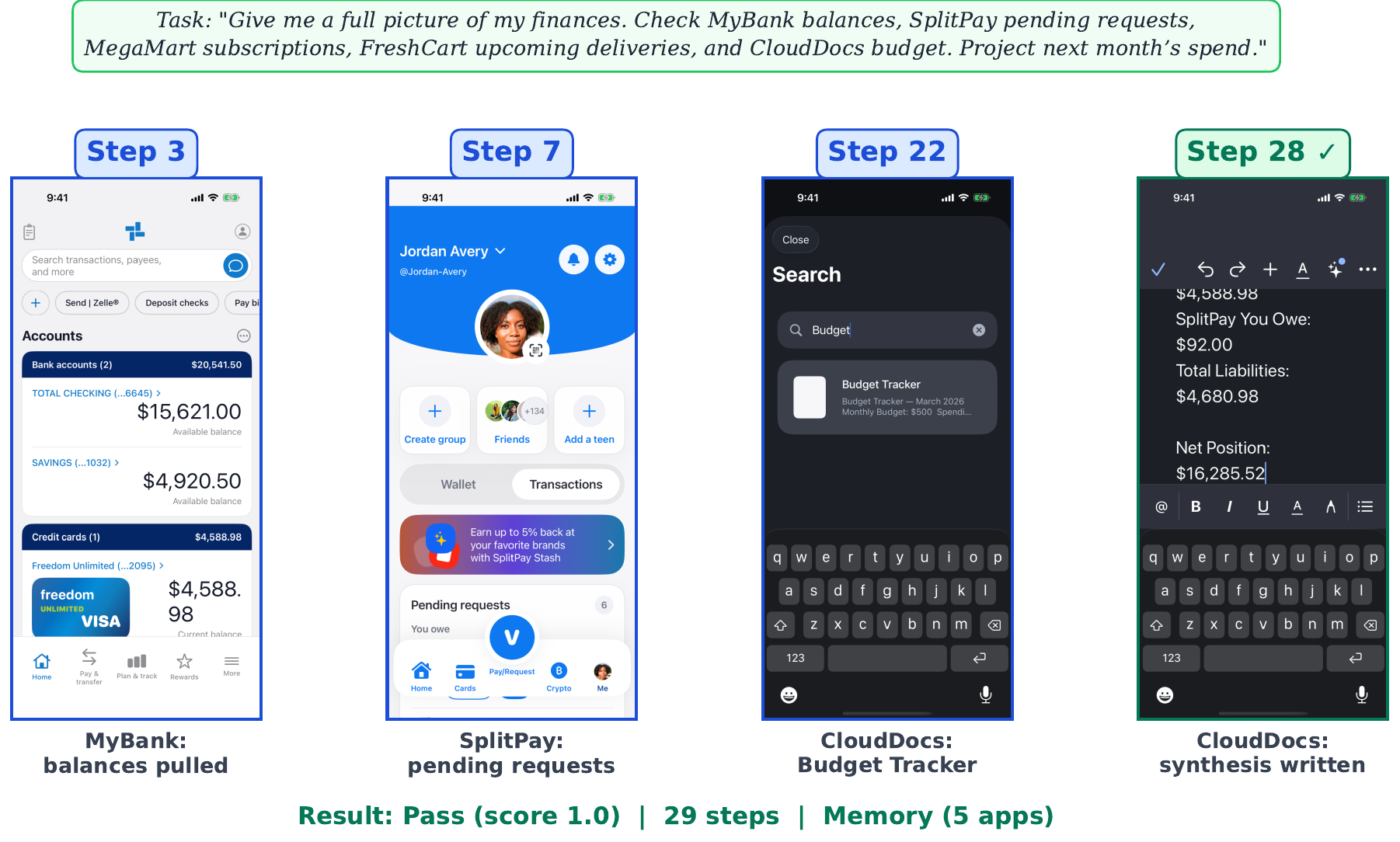}
\vspace{-0.1in}
\caption{\small Successful memory trajectory (Opus~4.6, vision+XML, 29 steps, score 1.0). For ``Give me a full picture of my finances,'' Opus pulls balances from MyBank, checks pending requests in SplitPay, opens the Budget Tracker in CloudDocs, and writes a synthesis spanning five apps in 29 steps.}
\label{fig:succ-memory-main}
\end{figure*}

Privileged vision+XML access helps the stronger frontier models (Fig.~\ref{fig:xml-advantage} and Appendix Fig.~\ref{fig:cua-gap}). Opus rises from 26\% to 52\% overall (+25.6 pp), Sonnet from 29\% to 47\% (+18.0 pp), and GPT-5.4 from 20\% to 40\% (+19.5 pp). Fig.~\ref{fig:xml-advantage} shows the gap on a multi-app QuickBite $\rightarrow$ TeamChat task. Vision-only Opus reaches checkout but gets stuck on a small payment-confirmation control and never opens TeamChat. Vision+XML Opus places the order, posts the deployment update, and checks \#general announcements in 22 steps. With vision+XML, Sonnet reaches 93\% on single-app tasks, while Opus leads on memory at 54\% and multi-app at 37\%. Multi-app tasks remain the hardest category. Fig.~\ref{fig:succ-memory-main} traces a successful memory trajectory, where Opus pulls balances from MyBank, checks SplitPay pending requests, then synthesizes a budget projection in CloudDocs. In vision-only mode, frontier models cluster between 20\% and 29\%. Sonnet (29\%) and Opus (26\%) lead through strong single-app numbers, while Gemini at 28\% is the most step-efficient at 21 steps per task versus 42--45 for Anthropic and OpenAI models. GPT-5.4 Mini and Qwen3.5 do not show the same gain from the extra accessibility-tree context, suggesting a capacity limit rather than a problem with the modality itself (see \S\ref{sec:analysis}).

\subsection{Analysis}
\label{sec:analysis}

\paragraph{Why does XML help so much?} Much of the vision-to-XML gap comes from ordinary iOS friction. Dense screens make coordinates hard to estimate, app switching can derail from the home screen, the accessibility tree can expose labels that are visually small or off-screen, and iOS has no universal back button. We therefore treat XML as privileged access, not just better text input. Across the 26 Opus tasks where vision-only fails (score $<$0.5) and vision+XML passes, $\sim$70\% include a home-screen or app-switching failure that \texttt{launch\_app} removes. The lift is largest on memory (Opus: 9\% $\rightarrow$ 54\%), where labels and values matter most. It also helps multi-app tasks once agents can launch and target apps reliably (Opus: 20\% $\rightarrow$ 37\%; Sonnet: 22\% $\rightarrow$ 35\%). Appendix~\ref{app:ios-patterns} quantifies two of these iOS-specific factors directly.

\paragraph{Smaller models struggle with the extra context.} More interface information is not always useful. GPT-5.4~Mini drops from 26\% vision-only to 16\% vision+XML, and 22 of the 35 tasks it solves vision-only become failures under XML. This is consistent with the added $\sim$3{,}100 tokens per step exceeding its effective context budget (Fig.~\ref{fig:mini-degrade}). Qwen3.5 35B-A3B degrades more sharply. XML takes it from 13\% to 11\% overall and from 7\% to 0\% on multi-app, with $\sim$50\% of its 119 XML failures dominated by action loops (Fig.~\ref{fig:qwen-loop}). With structured per-app MCP tools, pass rate rises from 12.8\% to 24.8\% and mean rubric score from 0.33 to 0.683 (Appendix~\ref{app:mcp-tools}).

\paragraph{Failure taxonomy.} We group the 422 frontier vision+XML failures into three modes. \textit{Budget exhausted} covers full 50-step runs and accounts for 51\% of failures. \textit{Gave up} covers early stops with score $<0.67$ and accounts for 26\%. \textit{Premature stops} covers early stops with score $\geq 0.67$ and accounts for 23\%. Budget exhaustion is most common on multi-app (55\%) and memory (52\%) tasks, while premature stopping is most common on single-app tasks (48\%). GPT-5.4~Mini gives up on 47\% of its failures. Qwen3.5 has a different profile, with $\sim$50\% of its 119 XML failures flagged as stuck-action loops under our $\geq$3-identical-action heuristic (Fig.~\ref{fig:qwen-loop}). Fig.~\ref{fig:failure-main} shows a representative budget-exhausted failure. Full breakdowns are in Appendix~\ref{app:failure-details}.

\begin{figure*}[t]
\vspace{-0.1in}
\centering
\includegraphics[width=0.82\textwidth,keepaspectratio]{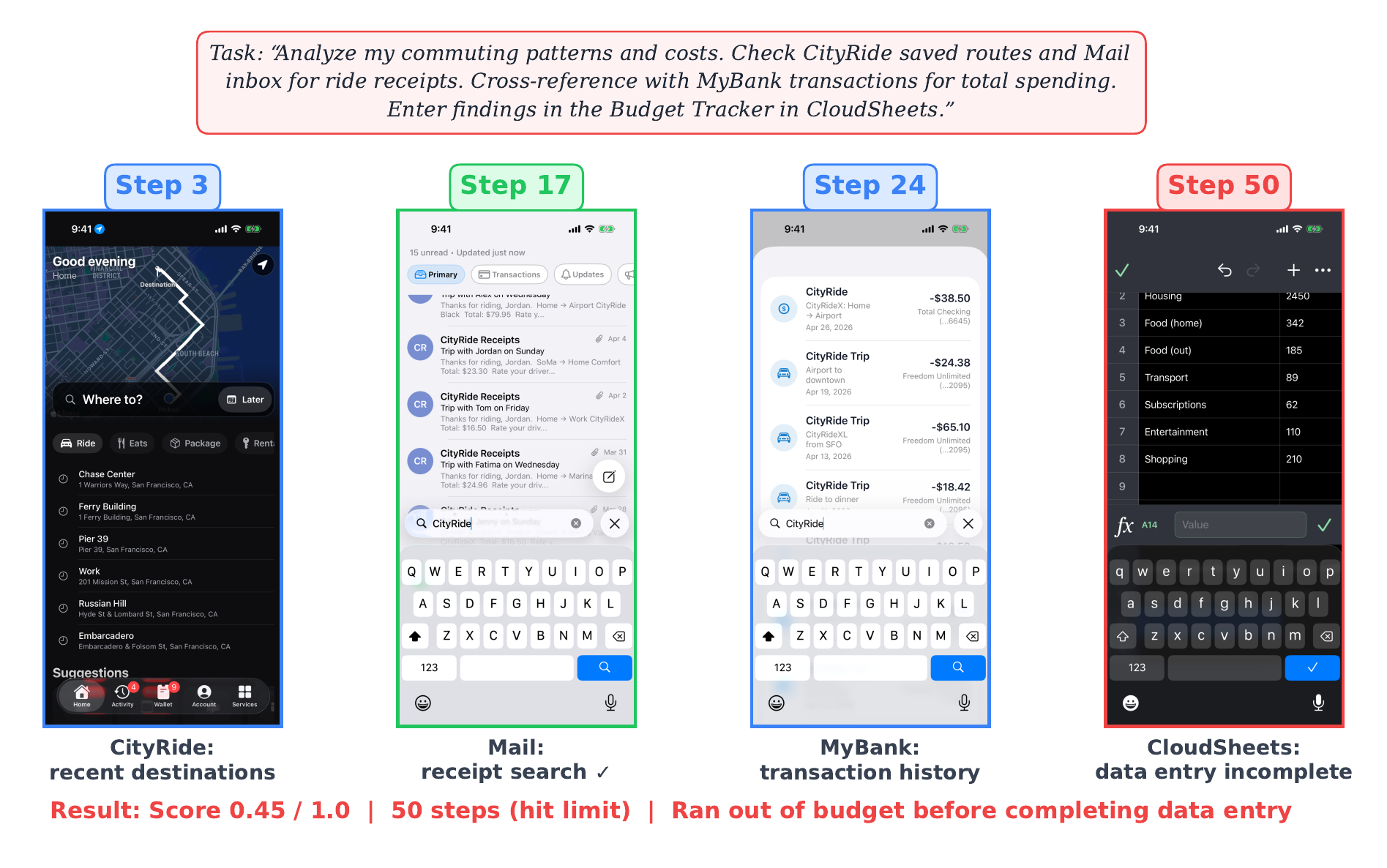}
\vspace{-0.1in}
\caption{\small A representative budget-exhausted failure (Opus~4.6, vision+XML, 50 steps, score 0.45). Opus explores CityRide (step~3), finds Mail receipts (step~17), reaches MyBank transactions (step~24), but exhausts the 50-step budget before completing data entry in CloudSheets. Budget-exhausted runs account for 51\% of frontier-model failures.}
\label{fig:failure-main}
\end{figure*}

\paragraph{Scaling and judge validation.} Step-budget curves (Fig.~\ref{fig:scaling}, appendix) show single-app tasks saturating by step~20, while multi-app tasks keep improving through step~40. Memory tasks vary more. Opus climbs from 17\% at step~30 to 54\% at step~50, whereas GPT-5.4~Mini plateaus at 16\% and Qwen3.5 reaches only 11\%. The trajectory judge agrees with human annotators on 128 Opus~4.6 trajectories at $\kappa$=0.77 task-level accuracy (89\%, F1=0.86) and $\kappa$=0.69 on rubric criteria (Pearson $r$=0.85). We find that cross-judge checks do not change the conclusions. Human-agreement details are in Appendix~\ref{app:human-agreement}.

\section{Conclusion}
\vspace{-0.05in}

\iosworld is an interactive native iOS benchmark with a persistent user identity across 26 apps. The best vision+XML configuration reaches 93\% on single-app tasks, but only 37\% on multi-app and 54\% on memory tasks. Qwen3.5 35B-A3B trails at 11\% overall. Even frontier models often run out of room, with 51\% of their failures exhausting the 50-step budget. Closing this gap will require stronger loop recovery, better action and visual grounding, and planning that is aware of the user's history. We release the code, environments, and an AWS runner at \url{https://iosworld.io}. Our environment and open-source code allow seamless addition of new tasks, personas, and apps. We believe that iOSWorld can provide a strong foundation for furthering the research on mobile agents and a shift towards emphasizing the personalization aspect of agents in deployment.

\newpage

\bibliographystyle{colm2026_conference}
\bibliography{colm2026_conference}

\appendix

\section*{Ethics Statement}

\paragraph{Synthetic data.} All data in \iosworld is entirely synthetic. The Jordan Avery persona is fictional, and no real user data was collected, processed, or used at any stage. Benchmark runs use deterministic seeded data and do not depend on real user accounts, real services, or external databases. We chose this design specifically to enable research on personalization and memory tasks without the privacy risks inherent in real user data.

\paragraph{Malicious Agents.} Phone agents capable of operating autonomously on a user's device carry significant dual-use risks. An agent with access to personal messaging, banking, ride-hailing, and email apps could be misused for surveillance, unauthorized transactions, social engineering, or data exfiltration. Even well-intentioned agents can cause harm through errors, such as sending a message to the wrong contact, making an unintended purchase, or leaking personal information across apps. The personalization and memory tasks in \iosworld are particularly sensitive because they require agents to reason about personal data, which is the most damaging if mishandled. We encourage researchers to develop agents with explicit user consent mechanisms and action confirmation for irreversible operations.

\paragraph{iOS access and reproducibility.} \iosworld requires macOS with Xcode to run the iOS Simulator, which limits reproducibility to researchers with access to Apple hardware. We release all source code, seed data, and evaluation scripts. We also release an AWS-runner deployment (EC2-managed Mac instances) so non-Mac researchers can submit task batches and receive the same evaluation bundle. The closed-source nature of iOS means the vision+XML modality relies on XCUITest, a developer tool unavailable to a deployed consumer agent. Vision-only numbers reflect deployed capability, while vision+XML represents an upper bound with privileged access.

\paragraph{Single-persona scope.} The release uses one fictional user (Jordan Avery) to keep personalization tasks ground-truth-verifiable. The persona-seeding pipeline, schema, task generator, and rubric framework are released so contributors can generate a comparable task suite for a new persona. Multi-persona evaluation is left to future work.

\paragraph{Accessibility and intended use.} Capable phone agents could improve accessibility for users with visual, motor, or cognitive impairments who find complex multi-step workflows difficult. \iosworld is a research benchmark for measuring progress in a controlled simulator. Results should not be read as readiness for deployment on real devices with real user data.

\section{LLM Disclosure}
\label{app:llm-disclosure}

We used large language models in several stages of this work. All drafting and structural decisions were made by human authors. Claude Code was used to polish prose, check grammar, and verify consistency with human-in-the-loop review. Figures and plots were generated programmatically via Claude Code from human-provided sketches, with the human author directing layout and content at every iteration. We used a multimodal LLM coding agent to perform high-level quantitative analysis, such as aggregating scores, computing pass rates, and to flag qualitative patterns in agent trajectories (e.g., identifying failure modes from screenshots). All flagged results were reviewed, verified, and synthesized into written analysis by human authors. The 26 iOS applications were built in SwiftUI using Claude Code as a coding assistant with human developers verifying correctness, and tasks and rubrics were generated by Claude Code then reviewed, refined, and manually executed end-to-end by human annotators. Finally, we use GPT-5.4 Mini as an LLM-as-a-judge evaluator, validated against human annotators ($\kappa$=0.77).

\section{Application Details and Dataset Statistics}

Table~\ref{tab:apps} lists all 26 applications. QuickChat (44 task references), Notes (41), CloudDocs (35), and Mail (29) are the most frequently involved apps.

\begin{table*}[t]
\centering
\small
\resizebox{\textwidth}{!}{%
\begin{tabular}{@{}llp{9.5cm}@{}}
\toprule
\textbf{App} & \textbf{Analogue} & \textbf{Key Seed Data and Features} \\
\midrule
MyBank & Chase & Checking \$15,621; Savings \$4,920; Credit \$4,589/\$8,000; 611 ledger transactions \\
SplitPay & Venmo & 17 core users, 24 suggestions, 80 transactions, \$650/mo rent from Arnav, 6 pending \\
QuickChat & WhatsApp & 54 contacts, 40 conversations (26 DMs + 14 groups) \\
TeamChat & Slack & 25 members, 11 channels (\#eng-mobile, \#launch-war-room), 8 DMs \\
Mail & Mail & 84 inbox, 10 sent, 1 draft, 80 archive; senders span all other apps \\
SkyTrip & Delta & Gold Medallion, 127K miles, 5 upcoming trips (SFO$\rightarrow$JFK, SEA, ORD, HNL, LHR) \\
CityRide & Uber & Home/Office saved, 79 trips, 20 drivers, 5 ride types \\
StayFinder & Airbnb & Upcoming: Catalina Island, Barcelona sail. Past: Big Sur, Lisbon, London \\
QuickBite & DoorDash & 82 restaurants, 5 saved addresses, 30+ past orders (Chipotle, Sweetgreen) \\
FreshCart & Instacart & 13 stores, 112 products, 35 orders, scheduled/active/delivered \\
MegaMart & Amazon & Prime member, 10 depts, 17 cart items, 51 saved items, 42 orders \\
TasteRank & Beli & 60 restaurants, 8 saved lists (Date Night, Seoul Guide, etc.) \\
DineSpot & OpenTable & 74 restaurants, 7 cities, outdoor/tasting/vegetarian filters \\
CalTrack & MyFitnessPal & 77+ foods, 2,450 cal goal, 170g protein target, weight 182 lbs \\
CloudDocs & Google Docs & Mobile Launch Plan, Compute Access Runbook, Agent Systems Review \\
CloudDrive & Google Drive & Product Strategy, Shared Assets, Archive folders \\
CloudSheets & Google Sheets & Budget Tracker, Experiment Results (multi-sheet) \\
CloudSlides & Google Slides & Conference Practice Deck, Agent Systems Talk \\
Notes & Apple Notes & 4 folders, 14 notes, 3 pinned (Shopping List, Wifi Passwords, Team Standup) \\
Cinephile & Letterboxd & Wishlists, Seenlist, custom lists, fan clubs \\
TicketBox & SeatGeek & 17 venues, 66 events, Taylor Swift Eras Tour, Hamilton \\
TrailBlaze & Strava & Bundled snapshot with 4 athletes, 3 routes, 6 activities, PRs (5K: 19:42), 2 clubs \\
ScoreZone & ESPN & Favorites: Lakers, Chiefs, Yankees; 40 seeded teams, scores \\
Weather & Weather & SF, NYC, Dallas; hourly and 10-day forecasts \\
Clock & Clock & Alarms, world clocks (Tokyo, London), stopwatch, timer \\
LockedIn & LinkedIn & Connections, job postings, professional profile \\
\bottomrule
\end{tabular}%
}
\caption{The 26 iOS applications in \iosworld with real-world analogues and seed data.}
\label{tab:apps}
\end{table*}

\paragraph{Per-app difficulty.} Table~\ref{tab:per-app} shows pass rate for Opus 4.6 (vision+XML) across all 26 apps. Cinephile is hardest (12\%, 8 references), followed by CloudDrive (14\%, 7 references), while CalTrack is easiest (65\%, 17 references). Mail (59\%, 29 references) and MyBank (59\%, 22 references) are also among the strongest. QuickChat remains challenging (20\%, 44 references) due to precise thread navigation across many tasks.

\begin{table}[t]
\centering
\small
\begin{tabular}{@{}lrrr@{}}
\toprule
\textbf{App} & \textbf{Pass} & \textbf{Total} & \textbf{Rate} \\
\midrule
CalTrack & 11 & 17 & 65\% \\
Mail & 17 & 29 & 59\% \\
MyBank & 13 & 22 & 59\% \\
ScoreZone & 5 & 9 & 56\% \\
TeamChat & 12 & 22 & 55\% \\
QuickBite & 9 & 17 & 53\% \\
MegaMart & 9 & 17 & 53\% \\
StayFinder & 5 & 10 & 50\% \\
CloudDocs & 17 & 35 & 49\% \\
CityRide & 8 & 17 & 47\% \\
TrailBlaze & 7 & 15 & 47\% \\
TicketBox & 7 & 16 & 44\% \\
Notes & 17 & 41 & 41\% \\
LockedIn & 4 & 10 & 40\% \\
SplitPay & 9 & 23 & 39\% \\
Clock & 6 & 16 & 38\% \\
CloudSlides & 3 & 8 & 38\% \\
SkyTrip & 4 & 11 & 36\% \\
FreshCart & 5 & 15 & 33\% \\
TasteRank & 4 & 12 & 33\% \\
Weather & 6 & 21 & 29\% \\
CloudSheets & 5 & 18 & 28\% \\
QuickChat & 9 & 44 & 20\% \\
DineSpot & 3 & 17 & 18\% \\
CloudDrive & 1 & 7 & 14\% \\
Cinephile & 1 & 8 & 12\% \\
\bottomrule
\end{tabular}
\caption{Per-app pass rate for Opus 4.6 (vision+XML) across all 26 apps.}
\label{tab:per-app}
\end{table}

\section{Rubric-Based Evaluation Details}
\label{app:rubric-details}

Each task is accompanied by a rubric, a list of independently verifiable criteria decomposing the objective into steps. The benchmark contains 1,123 rubric items across 133 tasks, ranging from 4 to 13 per task (mean 8.4). Multi-app tasks are the most rubric-dense at 9.4 items on average, reflecting the number of intermediate steps needed to coordinate across applications.

\paragraph{Rubric scores reveal partial progress.} Binary pass rates understate agent capability. Under vision+XML, the average rubric score (fraction of criteria satisfied) ranges from 29\% (Qwen3.5) to 81\% (Opus), meaning frontier agents satisfy a majority of criteria even on tasks they ultimately fail. The rubric perfect rate (all criteria satisfied) tracks the binary pass rate within about 0.8--2.3 percentage points, confirming internal consistency between the holistic judge and per-criterion evaluation.

\paragraph{Per-step evaluation.} We also evaluated a per-step variant in which the judge reviews each screenshot independently and we take the maximum across steps per criterion. It yields higher rubric scores (73--87\% average) but proved more lenient than the trajectory judge when validated against human annotators ($\kappa$=0.51--0.61 vs.\ 0.77 for the trajectory judge). Its mean rubric score is 0.83 versus 0.70 for humans, producing more than twice as many false-positive criteria (188 vs.\ 79). Since \iosworld tasks are relatively straightforward and compact (mean 8.4 criteria, max 50 steps), the trajectory judge provides sufficient discrimination without the added complexity of per-step evaluation. We use the trajectory-level judge throughout the main text.

\section{Failure Analysis Details}
\label{app:failure-details}

\begin{table}[t]
\centering
\small
\begin{tabular}{@{}lrrrr@{}}
\toprule
& \multicolumn{3}{c}{\textbf{Category (\%)}} & \\
\cmidrule(lr){2-4}
\textbf{Failure Mode} & {Single} & {Multi} & {Mem} & {All} \\
\midrule
Budget exhausted & 14 & 55 & 52 & 51 \\
Gave up          & 38 & 21 & 31 & 26 \\
Premature stop   & 48 & 24 & 18 & 23 \\
\bottomrule
\end{tabular}
\caption{Failure mode distribution under vision+XML across the five frontier models.}
\label{tab:failure-modes}
\end{table}

\begin{figure*}[t]
\centering
\includegraphics[width=0.92\textwidth,keepaspectratio]{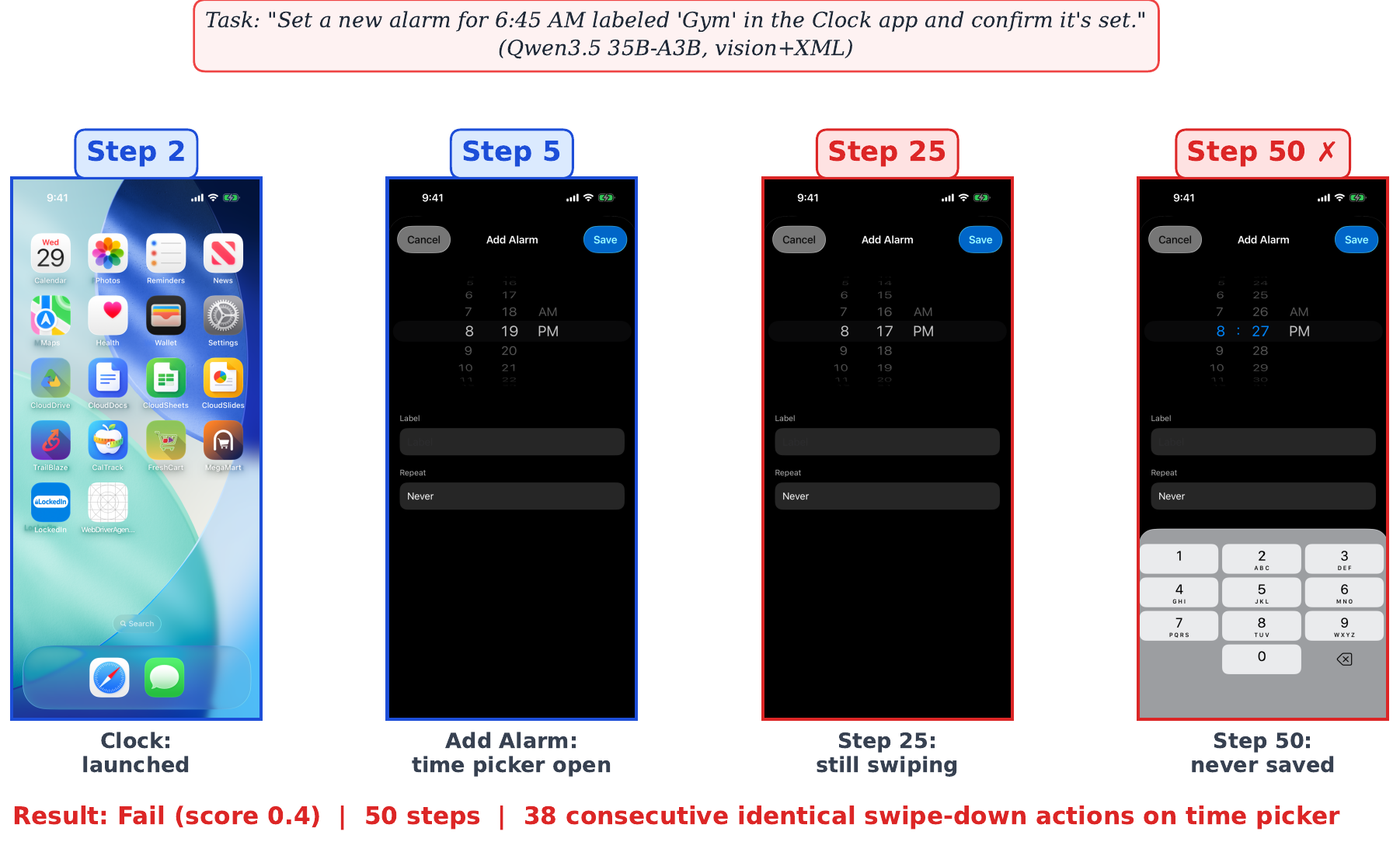}
\caption{\small Qwen3.5 35B-A3B (vision+XML) on a simple ``Set a 6:45 AM alarm labeled Gym'' single-app task. The agent reaches the Add Alarm screen by step 5 but then issues the \emph{same} swipe-down action on the time picker 38 consecutive times (steps 6--46), never adjusting to 6:45, never setting the label, and never tapping Save. The 50-step budget is exhausted on a task Opus and Sonnet both solve in 25 steps. Repeated-action loops account for $\sim$50\% of Qwen3.5's 119 XML failures.}
\label{fig:qwen-loop}
\end{figure*}

\paragraph{Methodology.} Each failed trajectory is assigned to exactly one of three mutually exclusive modes, derived from its step count and final rubric score. A run that stopped before the 50-step limit is a \textit{premature stop} if its rubric score is $\geq 0.67$ (it ended a largely correct trajectory too early) and \textit{gave up} otherwise. A run that used all 50 steps is \textit{budget exhausted}. We do not split budget-exhausted frontier runs into loop failures versus continued effort, because the two are hard to separate without human review. An automatic heuristic ($\geq$3 consecutive near-identical actions) over-flags benign repeated scrolls and near-coincident taps. For the open-source Qwen3.5 baseline the same heuristic is reliable, because its loops are blatant (e.g., 38 consecutive identical swipes; Fig.~\ref{fig:qwen-loop}). We report that loop share separately below.

\paragraph{Examples.} In a \textit{budget exhausted} case, Opus hits the 50-step limit on a commuting-patterns memory task with partial progress across CityRide, Mail, MyBank, and CloudSheets (mem-021, score 0.45; Fig.~\ref{fig:failure-main}). In a \textit{premature stop}, GPT-5.4 reports fare and ETA correctly but stops at the final ``Request'' button after 8 steps without confirming the booking (cityride-001, score 0.80). In a \textit{gave up} case, Opus abandons a game-day planning task after 44 steps with only partial progress across ScoreZone, QuickChat, and SplitPay (mem-020, score 0.65).

\paragraph{Open-source baseline (Qwen3.5 35B-A3B).} 119/133 vision+XML failures (10.5\% pass rate). The $\geq$3-consecutive-identical-actions loop heuristic is reliable here because Qwen's loops are extreme: 50\% of its XML failures (60/119) are flagged as stuck-action loops.

\input{appendix_mcp_tools_ablation}

\section{Additional Results Figures}

\begin{figure}[t]
\centering
\includegraphics[width=\columnwidth]{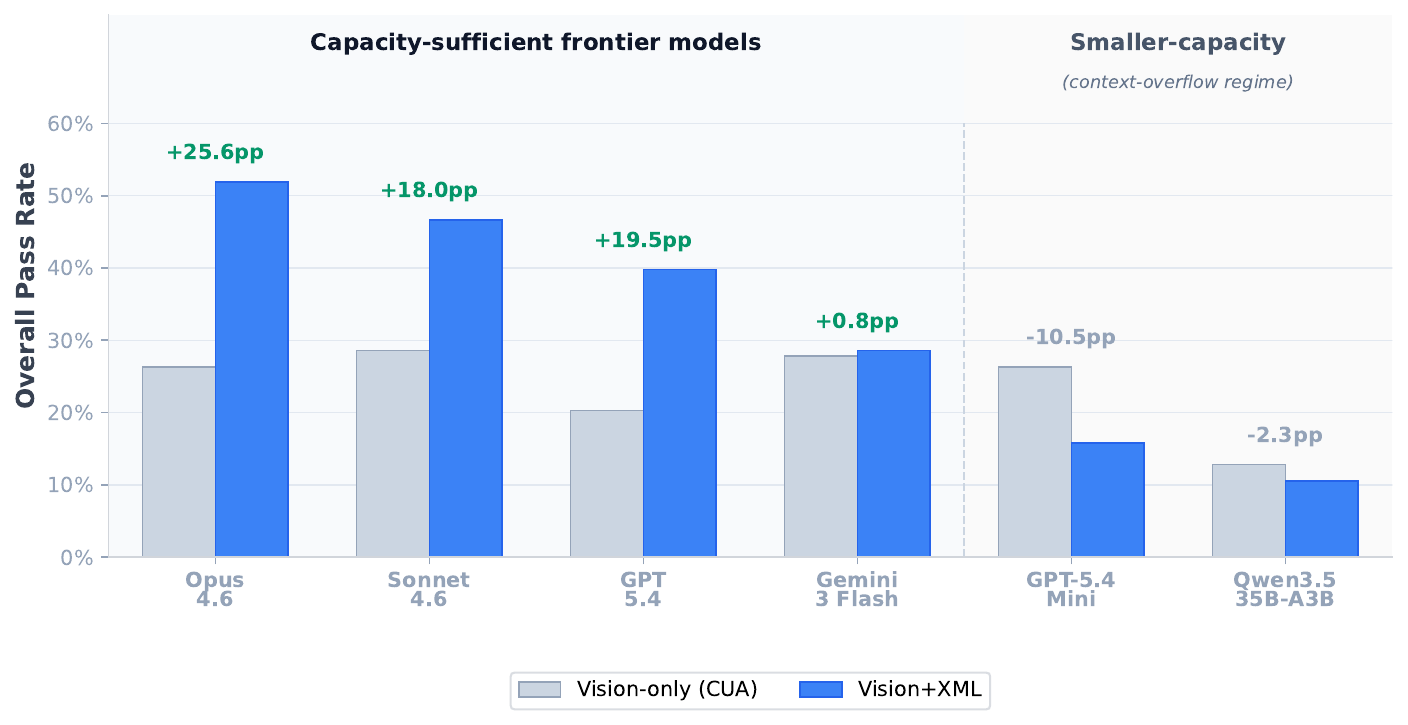}
\caption{Vision-only vs.\ vision+XML accuracy per model. Privileged vision+XML access improves the stronger frontier models (Opus +25.6 pp, Sonnet +18.0 pp, GPT-5.4 +19.5 pp, Gemini +0.8 pp). Smaller models (GPT-5.4 Mini, Qwen3.5 35B-A3B) do not benefit from the additional accessibility-tree input.}
\label{fig:cua-gap}
\end{figure}

\begin{figure*}[t]
\centering
\includegraphics[width=0.95\textwidth]{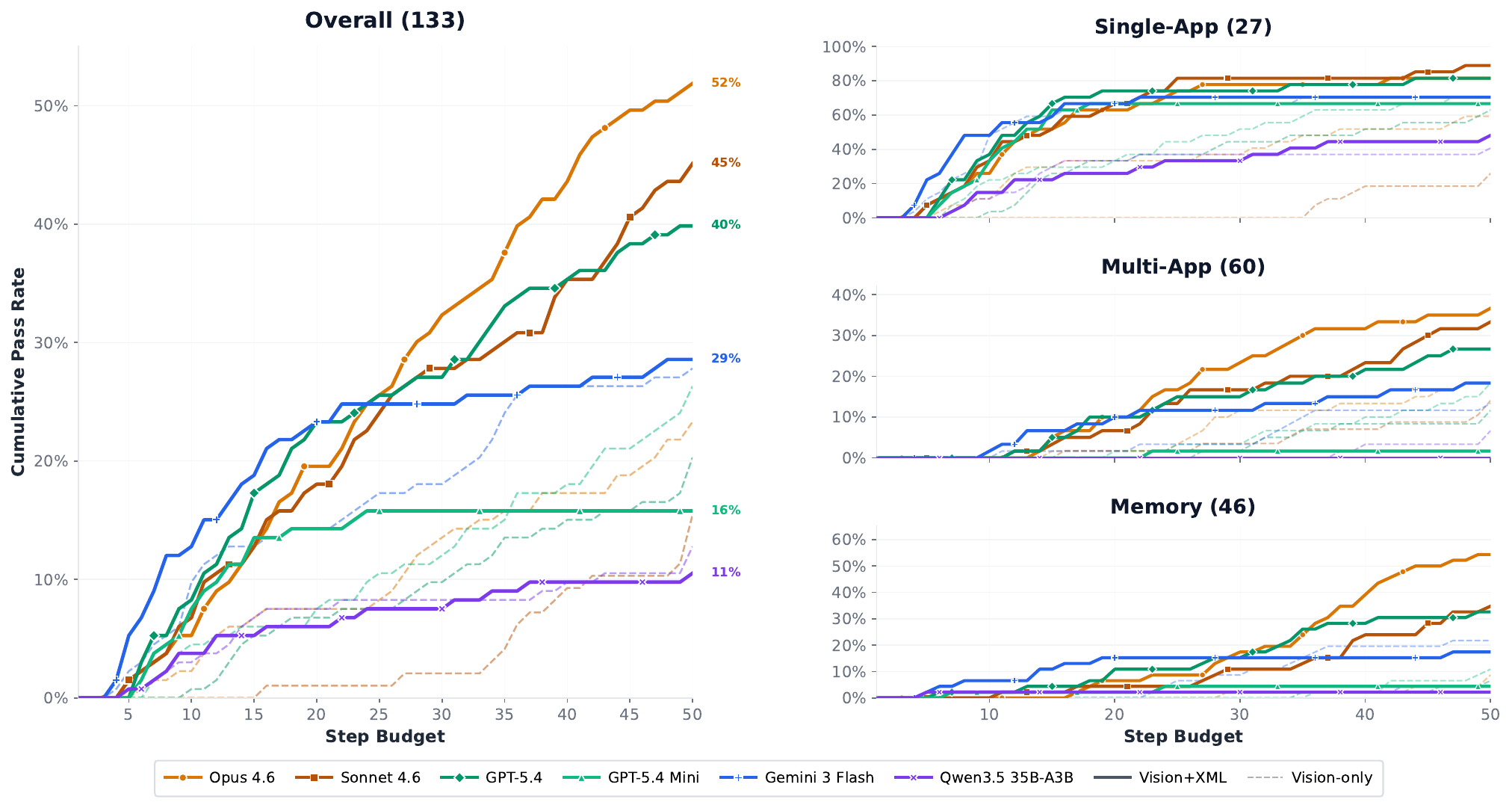}
\caption{Cumulative pass rate vs.\ step budget. \textbf{Left:} overall. \textbf{Right:} by task category. Solid: vision+XML; dashed: vision-only. Single-app saturates by step~20; multi-app scales through step~40; memory shows varied scaling with Opus climbing steeply past step~30.}
\label{fig:scaling}
\end{figure*}

\begin{figure*}[t]
\centering
\includegraphics[width=\textwidth]{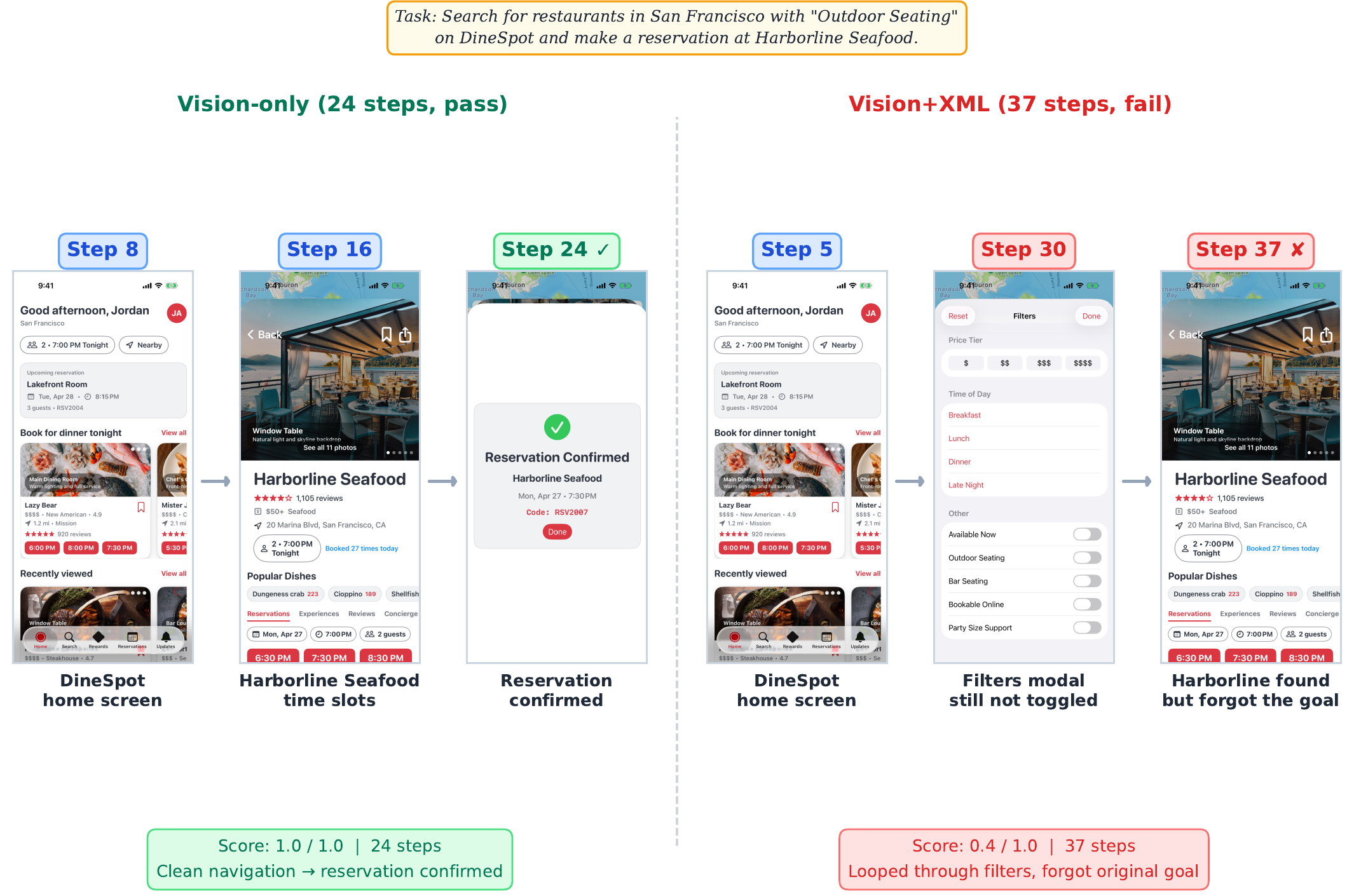}
\caption{\small GPT-5.4 Mini on the same DineSpot reservation task under both modalities. \textbf{Left}: vision-only navigates cleanly to a confirmed booking in 24 steps (score 1.0). \textbf{Right}: vision+XML loops through filter menus for 30 steps and ultimately forgets the original goal (score 0.4, 37 steps). The accessibility tree overwhelms the smaller model's context capacity.}
\label{fig:mini-degrade}
\end{figure*}

\begin{table}[t]
\centering
\small
\begin{tabular}{@{}lcccccc@{}}
\toprule
\textbf{Action} & \textbf{Opus} & \textbf{Sonnet} & \textbf{GPT-5.4} & \textbf{Mini} & \textbf{Gemini} & \textbf{Qwen3.5} \\
\midrule
tap\_xy & 62\% & 62\% & 55\% & 37\% & 54\% & 66\% \\
swipe & 21\% & 23\% & 12\% & 35\% & 10\% & 13\% \\
type & 9\% & 8\% & 18\% & 21\% & 17\% & 4\% \\
launch\_app & -- & -- & -- & -- & 18\% & 1\% \\
wait & 1\% & 1\% & 9\% & 3\% & $<$1\% & 1\% \\
home & 7\% & 7\% & 7\% & 3\% & $<$1\% & 16\% \\
\bottomrule
\end{tabular}
\caption{Action type distribution per model under vision+XML.}
\label{tab:actions-dist}
\end{table}

\clearpage

\section{App and Task Construction}

\paragraph{App creation.} Apps were created or adapted using Claude Code with a structured prompt specifying constraints (SwiftUI, deterministic seeded data, accessibility identifiers), workflows, data models, and seed quantities. Apps underwent iterative refinement and manual verification by human developers.

\paragraph{Task pipeline.} Stage 1: Claude Code generated tasks grounded in app source code and seed data. Stage 2: A Python pipeline normalized app names, rewrote tasks in first-person voice, and generated rubric criteria. Stage 3: Human annotators executed every task on the simulator.

\section{Prompts}

\paragraph{Agent system prompt.} All models receive the following iOS-specific instructions. Action names are adapted per provider (e.g., \texttt{left\_click} for Claude, \texttt{click} for OpenAI, \texttt{click\_at} for Gemini; see Tab.~\ref{tab:action-translation}). The version below is for Claude CU; the vision+XML variant appends the accessibility tree instructions at the end.

\begin{quote}
\small
\texttt{You are controlling an iOS Simulator (iPhone). This is a touch-screen mobile phone with NO mouse cursor, NO physical keyboard shortcuts, and NO right-click.}

\texttt{CURRENT STATE: You start on the iOS home screen. You must find and open apps yourself.}

\texttt{HOW TO OPEN APPS:}\\
\texttt{- Tap an app icon on the home screen if it is visible.}\\
\texttt{- To search for an app: swipe DOWN from the MIDDLE of the home screen to open Spotlight search, then type the app name and tap the result.}\\
\texttt{- Swipe left/right on the home screen to browse additional pages of apps.}

\texttt{TOUCH INTERACTIONS:}\\
\texttt{- Use `left\_click' for all touch/tap interactions (there is no mouse cursor).}\\
\texttt{- To type text, click a text field first, then use the `type' action to enter text.}\\
\texttt{- To scroll content, use the `scroll' action with delta\_x/delta\_y.}

\texttt{iOS-SPECIFIC BEHAVIOURS:}\\
\texttt{- HOME: Use key `Home', or swipe up from the very bottom of the screen.}\\
\texttt{- APP SWITCHER: Swipe up from the bottom and pause mid-screen.}\\
\texttt{- BACK NAVIGATION: Look for a back button (top-left) or swipe from the left edge.}\\
\texttt{- KEYBOARD DISMISS: Tap any area outside the text field.}

\texttt{COMPLETING THE TASK:}\\
\texttt{- When the task is complete, stop calling the computer tool and respond with a text summary.}\\
\texttt{- IMPORTANT: If the task asks you to find, check, look up, or report ANY information, you MUST include that exact information in your final text response.}

\texttt{ACCESSIBILITY TREE (appended in vision+XML mode only):}\\
\texttt{On each turn you will also receive a text accessibility tree of the current UI. The tree lists every element with its type, name/label, value, accessibility IDs (shown as id="..."), and centre coordinates.}

\texttt{IMPORTANT: The coordinates in the tree are in the SAME coordinate space as your action coordinates. You can use tree coordinates DIRECTLY as click/tap targets without any conversion or mapping.}

\texttt{How to use the two inputs together:}\\
\texttt{- The SCREENSHOT is ground truth for what is displayed on screen.}\\
\texttt{- The TREE provides precise element names and coordinates for targeting.}\\
\texttt{- Use tree coordinates DIRECTLY to click more precisely than visual estimation.}\\
\texttt{- If the screenshot and tree disagree (e.g. an element appears in the tree but not on screen), trust the screenshot; the element may be off-screen or obscured.}\\
\texttt{- Elements marked [hidden] are in the DOM but not rendered on screen.}\\
\texttt{- If you need to find elements not currently visible, try scrolling.}
\end{quote}

\paragraph{Trajectory-level evaluation.} The judge (GPT-5.4 Mini) receives the following prompt structure. Per-step screenshots are attached as images:

\begin{quote}
\small
\texttt{Goal: [task instruction]}

\texttt{You are evaluating whether an iOS agent successfully completed the above goal. The agent executed N steps. Full trajectory with per-step screenshots:}\\
\texttt{~~Step 1: [Screenshot 1 - before actions] Actions: [JSON]}\\
\texttt{~~Step 2: ...}\\
\texttt{~~[Screenshot N+1 - final state after all actions]}

\texttt{The agent’s final answer was: "[answer]"}

\texttt{N screenshots are attached, one per step showing the screen state before each action, plus one final screenshot showing the end state.}

\texttt{Evaluate each of the following rubric criteria individually:}\\
\texttt{~~1. [criterion]}\\
\texttt{~~2. ...}

\texttt{Respond with ONLY a JSON object (no code fences):}\\
\texttt{\{"success": true/false, "reasoning": "overall assessment", "rubric\_results": [\{"criterion": "...", "satisfied": true/false, "reasoning": "..."\}]\}}

\texttt{success=true means the goal is fully and completely achieved. For each rubric criterion, set satisfied=true only if there is clear evidence in the trajectory and screenshots that the criterion was met. For tasks that ask a question, evaluate whether the agent’s final answer is correct.}
\end{quote}

\paragraph{Per-step evaluation.} Each step is evaluated independently in a separate LLM call. The judge receives one screenshot, the agent’s action, and its reasoning. For the final step, the agent’s answer is also included:

\begin{quote}
\small
\texttt{Goal: [task instruction]}

\texttt{You are evaluating an iOS agent’s progress at step K. The attached screenshot shows the device screen after the agent’s action.}

\texttt{Agent’s action: [JSON]}\\
\texttt{Agent’s reasoning: [text]}\\
\texttt{Agent’s final answer: [text] ~~(last step only)}

\texttt{Which of the following rubric criteria are NOW satisfied based on the screenshot, the agent’s action, and its reasoning?}\\
\texttt{~~1. [criterion]}\\
\texttt{~~2. ...}

\texttt{Respond with ONLY a JSON object:}\\
\texttt{\{"satisfied": [list of criterion numbers that are satisfied]\}}\\
\texttt{Return an empty list if none are satisfied. Only mark a criterion satisfied if there is clear evidence from the screenshot AND the agent’s actions/reasoning. Do not infer satisfaction from ambiguous or partial evidence.}
\end{quote}

All steps are evaluated in parallel (up to 8 concurrent calls per task), and we take the max across steps per criterion. Once a criterion is satisfied at any step, it remains satisfied.

\paragraph{CUA action translation.} Each provider uses its own action vocabulary. We map all native actions to our unified iOS action schema (Table~\ref{tab:action-translation}):

\begin{table}[t]
\centering
\small
\begin{tabular}{@{}lllll@{}}
\toprule
\textbf{iOS Action} & \textbf{Claude CU} & \textbf{OpenAI CUA} & \textbf{Gemini CU} & \textbf{Qwen mobile\_use} \\
\midrule
\texttt{tap\_xy} & \texttt{left\_click} & \texttt{click} & \texttt{click\_at} & \texttt{click} \\
\texttt{tap\_xy} $\times$2 & \texttt{double\_click} & \texttt{double\_click} & -- & -- \\
\texttt{tap\_xy} $\times$3 & \texttt{triple\_click} & \texttt{triple\_click} & -- & -- \\
\texttt{type} & \texttt{type} & \texttt{type} & \texttt{type\_text\_at} & \texttt{type} \\
\texttt{type} (keys) & \texttt{key} & \texttt{keypress} & \texttt{key\_combination} & \texttt{system\_button} \\
\texttt{swipe} & \texttt{scroll} & \texttt{scroll} & \texttt{scroll\_at} & \texttt{swipe} \\
\texttt{swipe} & -- & -- & \texttt{scroll\_document} & -- \\
\texttt{swipe} & \texttt{left\_click\_drag} & \texttt{drag} & \texttt{drag\_and\_drop} & -- \\
\texttt{home} & \texttt{key} Home & \texttt{keypress} Home & \texttt{go\_home} & \texttt{system\_button} Home \\
\texttt{wait} & \texttt{wait} & \texttt{wait} & \texttt{wait\_5\_seconds} & \texttt{wait} \\
\texttt{launch\_app} & -- & -- & \texttt{open\_app} & -- \\
\texttt{hover} & -- & -- & \texttt{long\_press\_at} & \texttt{long\_press} \\
\texttt{open\_url} & -- & -- & \texttt{open\_url} & -- \\
\texttt{stop} & (text) & (text) & (text) & \texttt{terminate} / \texttt{answer} \\
\bottomrule
\end{tabular}

\caption{Action translation from provider-native vocabularies to our iOS action schema. Scroll direction is inverted for all providers (scroll down = swipe up on touchscreen). Claude and OpenAI output pixel coordinates scaled to 0--1000. Gemini outputs 0--999 directly, and Qwen mobile\_use outputs 0--999 under the cookbook contract. Actions marked ``--'' are not available for that provider. Qwen mobile\_use follows the official Qwen3-VL mobile-agent cookbook tool schema.}

\label{tab:action-translation}
\end{table}

\section{Example Trajectories}
\label{app:failure}

\begin{figure*}[t]
\centering
\includegraphics[width=\textwidth]{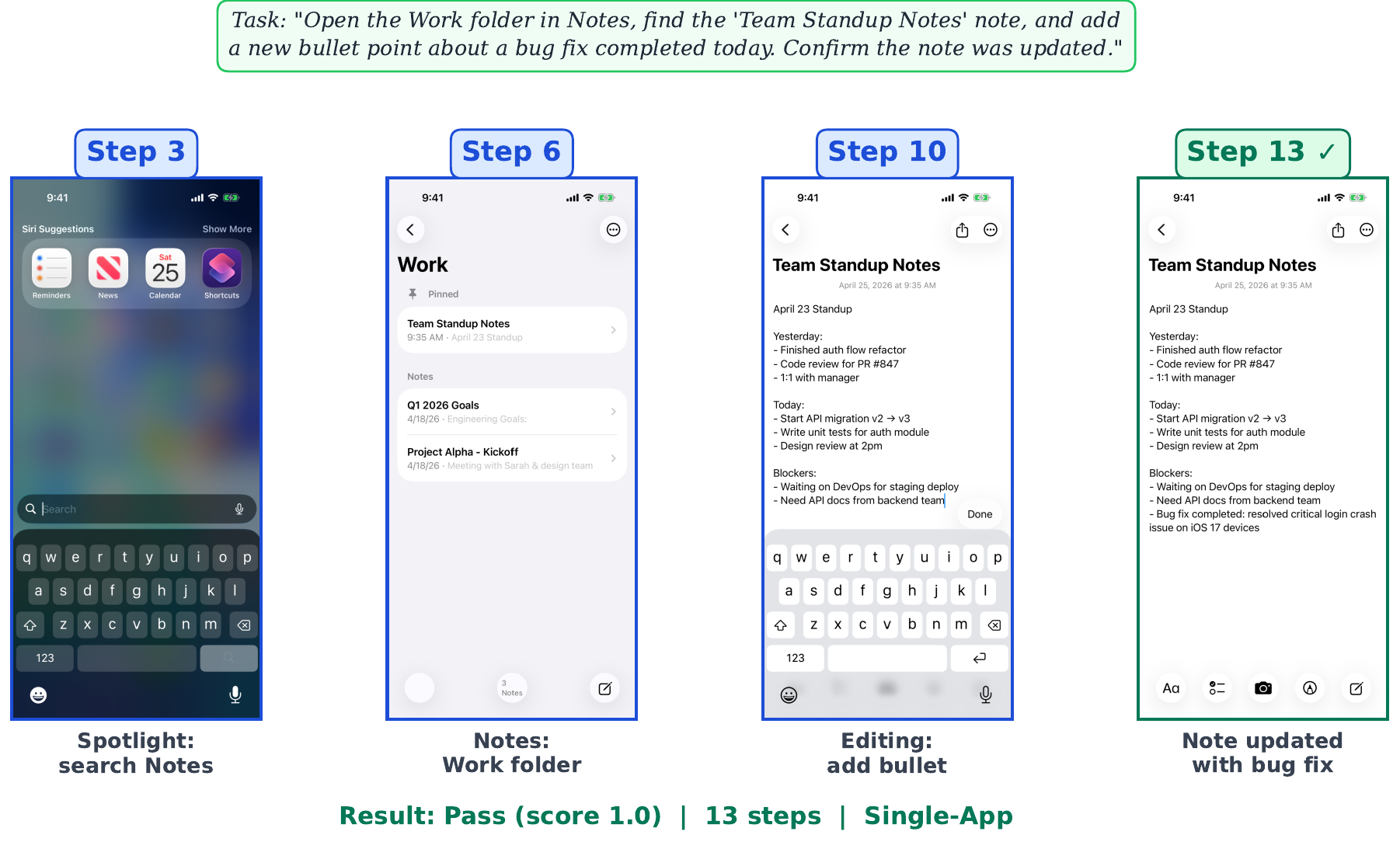}
\caption{Successful single-app Notes task. Team Standup Notes, add bug-fix bullet (Sonnet 4.6 CUA, 13 steps, score 1.0).}
\label{fig:succ-single}
\end{figure*}

\begin{figure*}[t]
\centering
\includegraphics[width=\textwidth]{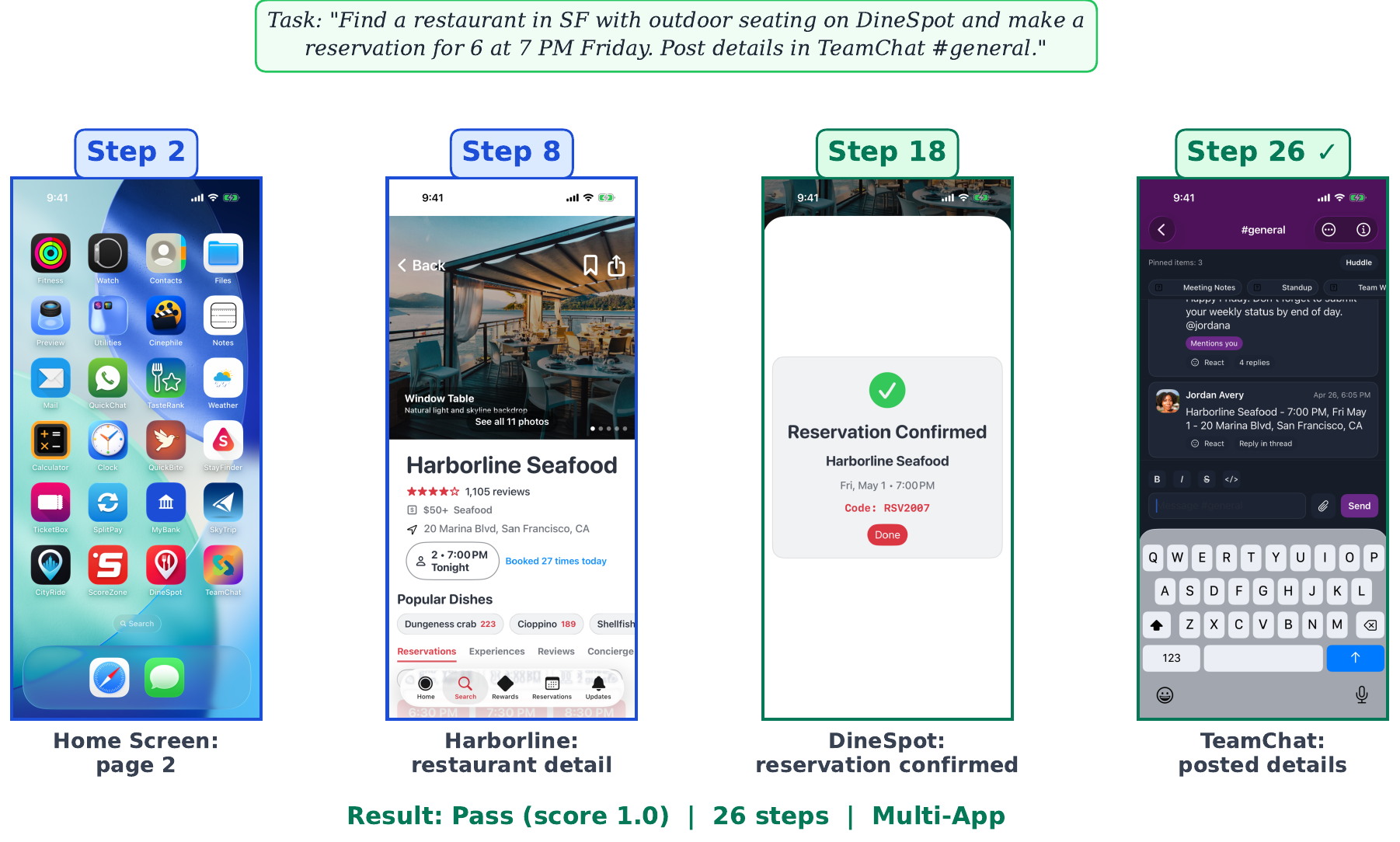}
\caption{Successful multi-app DineSpot $\rightarrow$ TeamChat task (Opus 4.6, vision+XML, 26 steps).}
\label{fig:succ-multi}
\end{figure*}

\begin{figure*}[t]
\centering
\includegraphics[width=\textwidth]{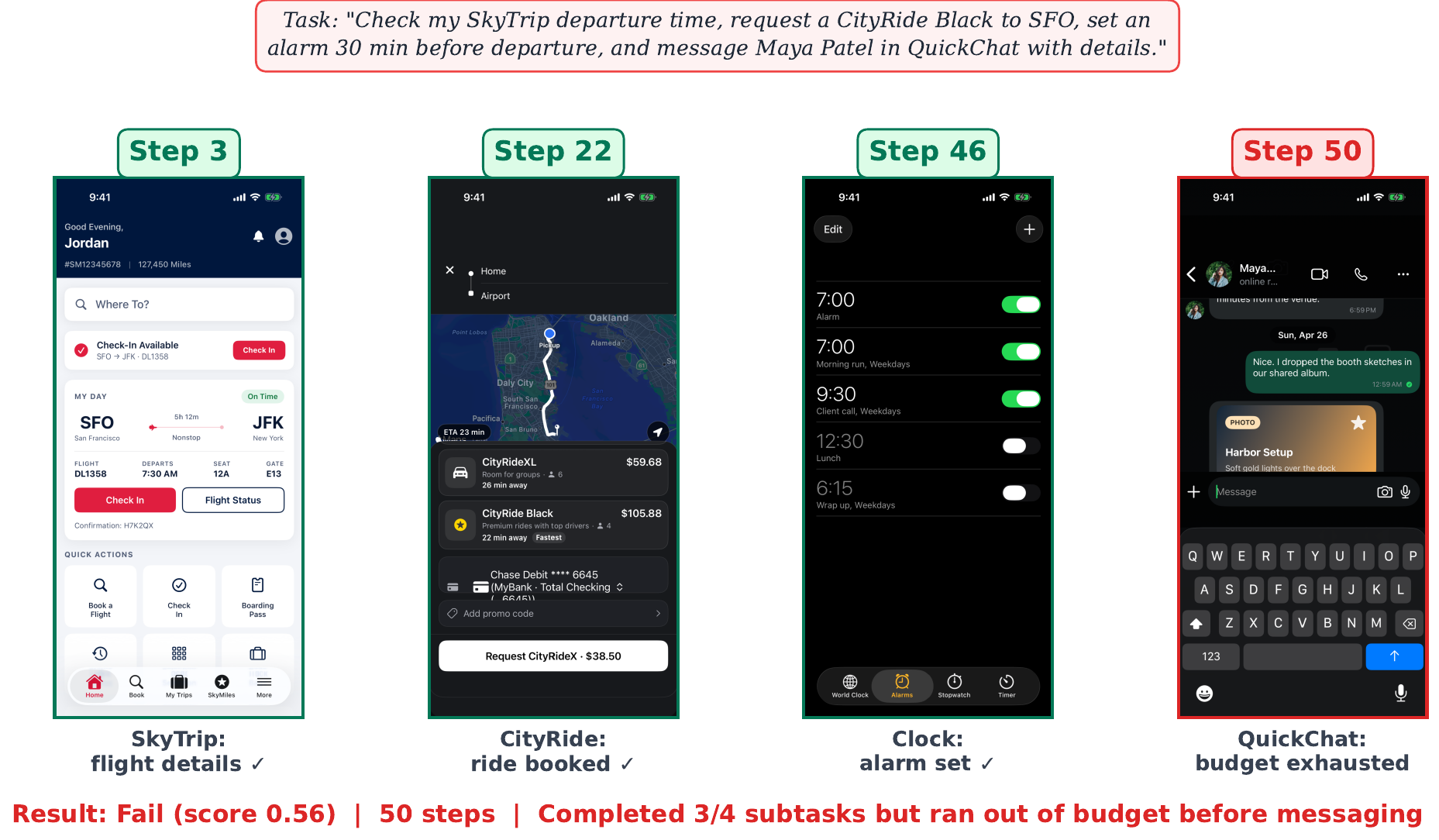}
\caption{Failed multi-app SkyTrip $\rightarrow$ CityRide $\rightarrow$ Clock $\rightarrow$ QuickChat task. The agent completed 3/4 subtasks but ran out of budget before messaging (50 steps, score 0.56).}
\label{fig:fail-multi}
\end{figure*}

\begin{figure*}[t]
\centering
\includegraphics[width=\textwidth]{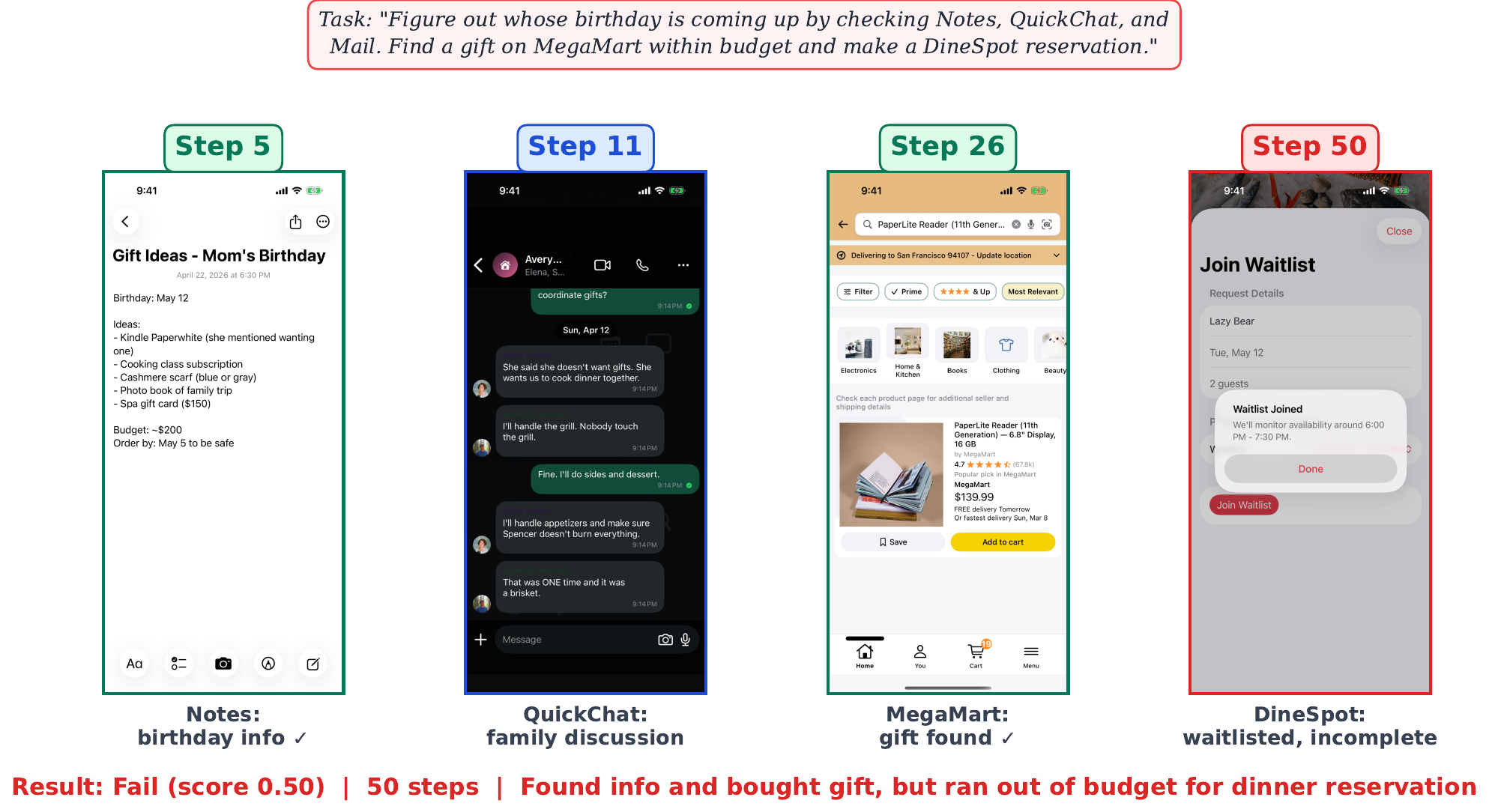}
\caption{Failed memory Notes $\rightarrow$ QuickChat $\rightarrow$ MegaMart $\rightarrow$ DineSpot task. The agent found birthday info and bought a gift but ran out of budget before the dinner reservation (50 steps, score 0.50).}
\label{fig:fail-memory}
\end{figure*}
\clearpage
\section{Human Agreement}
\label{app:human-agreement}

To validate the automated evaluation pipeline, we collect human annotations on 128 trajectories from the Opus~4.6 vision+XML configuration, spanning all three task categories. Four annotators each reviewed a subset of trajectories and graded every rubric criterion as \textit{pass} or \textit{fail}, along with an overall binary success judgment. We compare these human judgments against two automated judges. The trajectory-level judge sees the full action trace. The per-step judge evaluates each screenshot independently and takes the max across steps.

Table~\ref{tab:human-agreement-full} reports task-level and rubric-criterion agreement, plus the per-step judge comparison.

\begin{table}[t]
\centering
\small
\begin{tabular}{@{}lcccc@{}}
\toprule
\textbf{Comparison} & \textbf{Cohen's $\kappa$} & \textbf{F1} & \textbf{Acc.} & \textbf{$\rho$} \\
\midrule
\multicolumn{5}{@{}l}{\textit{Human vs.\ Trajectory Judge}} \\
\quad Binary task success  & 0.77 & 0.86 & 0.89 & 0.77 \\
\quad Rubric criteria ($n$=1,094) & 0.69 & 0.90 & 0.86 & 0.69 \\
\midrule
\multicolumn{5}{@{}l}{\textit{Human vs.\ Per-Step Judge}} \\
\quad Binary task success  & 0.61 & 0.79 & 0.80 & 0.63 \\
\quad Rubric criteria ($n$=1,094) & 0.51 & 0.87 & 0.81 & 0.56 \\
\midrule
\multicolumn{5}{@{}l}{\textit{Trajectory Judge vs.\ Per-Step Judge}} \\
\quad Binary task success  & 0.55 & 0.75 & 0.77 & 0.57 \\
\quad Rubric criteria ($n$=1,094) & 0.55 & 0.88 & 0.83 & 0.59 \\
\bottomrule
\end{tabular}
\caption{Extended agreement between human annotators and automatic judges on 128 trajectories (1,094 rubric criteria), including judge-vs-judge comparison. $\rho$ denotes Spearman correlation.}
\label{tab:human-agreement-full}
\end{table}

The trajectory-level LLM judge achieves substantial agreement with human annotators across all metrics. At the task level, binary success judgments agree 89\% of the time ($\kappa$=0.77, F1=0.86). At the rubric-criterion level, the judge correctly classifies 86\% of individual criteria ($\kappa$=0.69, F1=0.90). The rubric-level $\kappa$ is lower despite comparable accuracy because Cohen's $\kappa$ is sensitive to marginal distributions. Since 67\% of rubric criteria are satisfied, expected chance agreement is inflated and $\kappa$ is mechanically lower. Accuracy and F1 are more directly interpretable here. Continuous rubric scores are highly correlated, with Pearson $r$=0.85 and Spearman $\rho$=0.86 between human and trajectory-judge rubric fractions. Mean absolute error is 0.10. The per-step parallel judge shows lower agreement ($\kappa$=0.51--0.61), mainly because it is more lenient. Its mean rubric score is 0.83 versus 0.70 for humans, producing 188 false-positive criteria compared to 79 for the trajectory judge. This matches the design difference. Per-step evaluation marks a criterion satisfied if \emph{any} single screenshot shows evidence, which can overcount partial progress.

At the criterion level, the 148 disagreements between humans and the trajectory judge split into 79 false positives (LLM too lenient) and 69 false negatives (LLM too strict), indicating no strong bias in either direction.

\paragraph{Per-annotator analysis.} Table~\ref{tab:per-annotator} breaks down agreement by annotator. Four annotators each graded 26--47 trajectories. Task-level $\kappa$ ranges from 0.64 to 0.92, while rubric-level $\kappa$ is more tightly clustered (0.67--0.72), suggesting that per-criterion judgments are more consistent across annotators than holistic task-level judgments. The annotator with the lowest task-level $\kappa$ (0.64, 47~tasks) has the highest rubric-level $\kappa$ (0.72). This suggests that task-level disagreements come from borderline cases where most but not all criteria are satisfied, rather than from fundamentally different rubric interpretations. Human pass rates are consistent with the LLM judge across all annotators (36--50\% human vs.\ 36--46\% LLM), with no annotator showing a clear leniency or strictness bias. 
\clearpage
\begin{table}[t]
\centering
\small
\begin{tabular}{@{}lrcccc@{}}
\toprule
\textbf{Annotator} & \textbf{Tasks} & \textbf{$\kappa_{\text{task}}$} & \textbf{Acc.} & \textbf{$\kappa_{\text{rubric}}$} & \textbf{Criteria} \\
\midrule
A & 47 & 0.64 & 0.83 & 0.72 & 397 \\
B & 28 & 0.84 & 0.93 & 0.67 & 248 \\
C & 27 & 0.92 & 0.96 & 0.67 & 232 \\
D & 26 & 0.77 & 0.88 & 0.69 & 217 \\
\bottomrule
\end{tabular}
\caption{Per-annotator agreement with the trajectory judge. $\kappa_{\text{task}}$: Cohen's kappa on binary task success. $\kappa_{\text{rubric}}$: Cohen's kappa on individual rubric criteria. All annotators show substantial agreement ($\kappa \geq 0.64$).}
\label{tab:per-annotator}
\end{table}

\paragraph{Cross-judge robustness.} We re-scored all 128 validated Opus~4.6 vision+XML trajectories with five alternate judges (Table~\ref{tab:cross-judge}). The larger GPT-5.4 is the worst judge at $\kappa$=0.51 and over-rejects (1 FP vs.\ 27 FN against human, opposite of every other judge). Pairwise judge agreement sits in [0.74, 0.90] except GPT-5.4, which sits in [0.53, 0.74] against every other judge. 

\begin{table}[t]
\centering
\small
\begin{tabular}{@{}llcc@{}}
\toprule
\textbf{Judge model} & \textbf{Provider} & \textbf{$\kappa$ vs.\ human} & \textbf{Pass rate} \\
\midrule
Gemini 3 Flash & Google & \textbf{0.79} & 35.2\% \\
GPT-5.4 Mini (published) & OpenAI & 0.77 & 39.1\% \\
GPT-5 mini & OpenAI & 0.72 & 35.2\% \\
Claude Opus 4.6 & Anthropic & 0.68 & 30.5\% \\
Claude Sonnet 4.6 & Anthropic & 0.67 & 35.9\% \\
GPT-5.4 (full) & OpenAI & 0.51 & 20.3\% \\
\bottomrule
\end{tabular}
\caption{Cross-judge agreement with human annotators on the 128 validated Opus~4.6 vision+XML trajectories. Larger or in-family judges do not improve $\kappa$; GPT-5.4 (full) is an outlier due to systematic over-rejection.}
\label{tab:cross-judge}
\end{table}

\paragraph{Where the judge makes mistakes.} At the task level, disagreements are direction-dependent by category (Table~\ref{tab:judge-by-cat-task}). The judge over-accepts on single-app (3~FP, 0~FN), over-rejects on multi-app (0~FP, 5~FN), and is balanced on memory. At the criterion level (Table~\ref{tab:judge-by-criterion}), the judge is perfectly reliable on observable atomic actions (taps and swipes at 0\% error) and least reliable on semantic and report criteria such as ``did the agent correctly summarize X?'' (13 to 16\% error). These are also the criteria where humans have the most interpretive, subjective room.

\begin{table}[t]
\centering
\small
\begin{tabular}{@{}lcccc@{}}
\toprule
\textbf{Category} & \textbf{N} & \textbf{Over-accepts (FP)} & \textbf{Over-rejects (FN)} & \textbf{Agreement} \\
\midrule
Single-app & 26 & 3 & 0 & 88\% \\
Multi-app  & 58 & 0 & 5 & 91\% \\
Memory     & 44 & 3 & 3 & 86\% \\
\bottomrule
\end{tabular}
\caption{Task-level human--judge disagreement by task category (N=128).}
\label{tab:judge-by-cat-task}
\end{table}

\begin{table}[t]
\centering
\small
\begin{tabular}{@{}lrr@{}}
\toprule
\textbf{Criterion type} & \textbf{N} & \textbf{Error \%} \\
\midrule
tap / select / click & 27 & \textbf{0.0\%} \\
scroll / swipe & 2 & 0.0\% \\
type / enter input & 19 & 10.5\% \\
open / launch app & 270 & 11.5\% \\
send / submit / confirm & 127 & 11.8\% \\
search / filter & 23 & 13.0\% \\
identify / verify & 158 & 13.9\% \\
report / answer & 166 & 15.1\% \\
other (semantic) & 289 & \textbf{15.6\%} \\
\midrule
\textbf{TOTAL} & \textbf{1{,}090} & \textbf{13.2\%} \\
\bottomrule
\end{tabular}
\caption{Criterion-level human--judge disagreement by criterion type. The judge is perfectly reliable on observable atomic actions and least reliable on semantic / report criteria.}
\label{tab:judge-by-criterion}
\end{table}

\paragraph{Generalization beyond Opus.} We collected human annotations on 64 additional trajectories (32 Gemini~3~Flash and 32 GPT-5.4~Mini) through the same web tool (Table~\ref{tab:human-extended}). Agreement stays moderate to substantial on both new families ($\kappa$=0.49 and 0.60). The lower Gemini $\kappa$ partly reflects smaller N (32 vs.\ 128) and the lack of the multi-annotator calibration used on the original set.

\begin{table}[t]
\centering
\small
\begin{tabular}{@{}lrcccc@{}}
\toprule
\textbf{Agent family} & \textbf{N} & \textbf{Human pass} & \textbf{Judge pass} & \textbf{Agreement} & \textbf{Cohen $\kappa$} \\
\midrule
Anthropic (Opus 4.6) -- original   & 128 & 35.2\% & 39.1\% & 89\% & 0.77 \\
Google (Gemini 3 Flash) -- new     & 32  & 46.9\% & 40.6\% & 75\% & 0.49 \\
OpenAI (GPT-5.4 Mini) -- new       & 32  & 40.6\% & 34.4\% & 81\% & 0.60 \\
\bottomrule
\end{tabular}
\caption{Human--judge agreement extended to two non-Opus agent families. The judge generalizes beyond Opus; smaller N and single-annotator labeling explain part of the lower $\kappa$.}
\label{tab:human-extended}
\end{table}

\paragraph{Same-family bias check.} The published judge shares a provider with two evaluated agents (GPT-5.4 and GPT-5.4~Mini). We ran the out-of-family Gemini judge on the subset of GPT trajectories available for this bias audit (Table~\ref{tab:gemini-on-gpt}). This table is a cross-judge rescoring check, not the release aggregate in Table~\ref{tab:main-results}. Gemini passes GPT trajectories at a comparable or higher rate than the in-family published judge in every audited cell (mean delta +3.0~pp), which is opposite of what in-family inflation would predict. On the 32 human-validated GPT-5.4~Mini trajectories, Gemini also agrees with humans slightly better than the in-family judge ($\kappa$=0.66 vs.\ 0.60).

\begin{table}[t]
\centering
\small
\begin{tabular}{@{}llrccc@{}}
\toprule
\textbf{Agent} & \textbf{Mode} & \textbf{N} & \textbf{Published judge pass} & \textbf{Gemini judge pass} & \textbf{$\Delta$ (Gem $-$ Pub)} \\
\midrule
GPT-5.4       & vision & 128 & 12.5\% & 14.8\% & +2.3 pp \\
GPT-5.4       & xml    & 127 & 37.0\% & 44.1\% & +7.1 pp \\
GPT-5.4 Mini  & vision & 128 & 15.6\% & 17.2\% & +1.6 pp \\
GPT-5.4 Mini  & xml    & 128 & 33.6\% & 34.4\% & +0.8 pp \\
\bottomrule
\end{tabular}
\caption{Same-family judge bias check on the audited subset of OpenAI trajectories. These subset pass rates are cross-judge rescoring rates, not the release aggregates in Table~\ref{tab:main-results}. The out-of-family Gemini judge is comparable to or more lenient than the in-family published judge in every cell, opposite of what in-family inflation would predict.}
\label{tab:gemini-on-gpt}
\end{table}

\paragraph{Bootstrap stability.} A task-level bootstrap (5{,}000 resamples) on the Opus~4.6 vision+XML cell gives 95\% CIs of $\pm$8.3~pp Overall (mean 51.9\%), $\pm$11.7~pp Multi-app, $\pm$14.1~pp Memory, and $\pm$14.8~pp Single-app. The wider per-category CIs on the smaller subsets make multi-app and memory the more discriminative axes. Aggregate model ordering (Opus best Overall, Sonnet best Single-app, Gemini most step-efficient) is stable across resamples.

\section{iOS-Specific Interaction Patterns}
\label{app:ios-patterns}

We quantify two of the iOS-specific factors mentioned in \S\ref{sec:analysis} directly from the action trace.

\paragraph{Coordinate-grounding miss rate (vision-only).} A tap\_xy followed by another tap\_xy within 60~px on the next step is a strong proxy for a missed target followed by a retry. We compute this over all 133 vision-only trajectories per model, using the first tap in each step for tasks with multiple emitted actions (Table~\ref{tab:coord-miss}). Opus reaches 10.1\%, Sonnet 10.2\%, GPT-5.4 12.3\%, and GPT-5.4~Mini 10.3\%. Gemini is lower at 5.3\%, but its vision-only runs are shorter and leave fewer taps than the other frontier runs (1,753 vs.\ 3,102--4,384). Under XCUITest accessibility-ID taps the rate drops to $\approx$0\%, which isolates visual grounding on iOS-sized touch targets as a bottleneck distinct from reasoning.

\begin{table}[t]
\centering
\small
\begin{tabular}{@{}lrrr@{}}
\toprule
\textbf{Model} & \textbf{Total taps} & \textbf{Near re-taps} & \textbf{Miss rate} \\
\midrule
Claude Opus 4.6     & 3{,}102 & 314 & 10.1\% \\
Claude Sonnet 4.6   & 3{,}236 & 329 & 10.2\% \\
GPT-5.4             & 4{,}384 & 541 & \textbf{12.3\%} \\
GPT-5.4 Mini        & 3{,}792 & 392 & 10.3\% \\
Gemini 3 Flash      & 1{,}753 & 93  & 5.3\% \\
\bottomrule
\end{tabular}
\caption{Vision-only coordinate-grounding miss rate over all 133 tasks per model. Under XCUITest accessibility-ID taps the rate drops to $\approx$0\% across models.}
\label{tab:coord-miss}
\end{table}

\paragraph{Edge-swipe back-navigation under-use.} iOS has no hardware back button. Back navigation requires a left-edge rightward swipe or an in-app chevron. Across 12{,}255 frontier-model swipes, only 133 (1.1\%) are left-edge rightward swipes. GPT models use them somewhat more often in vision-only mode (2.7\% for GPT-5.4 and 2.1\% for GPT-5.4 Mini), but Claude and Gemini also use them rarely: at most 1.6\% in vision-only and 1.2\% in vision+XML. This points to broad under-use of the iOS gesture rather than a provider-specific behavior.

\end{document}

%% file: appendix_mcp_tools_ablation.tex

\section{MCP Tools Ablation}
\label{app:mcp-tools}

The main results attribute much of the vision-only/vision+XML gap to interface rather than reasoning. Precise element targeting and \texttt{launch\_app} remove failure modes introduced by coordinate estimation (\S\ref{sec:analysis}). We test that interpretation on the open-source baseline by holding the model, task set, judge, and 50-step budget fixed while varying only the action interface. Screenshots remain available, but the agent receives a structured per-app tool layer instead of the 7-action cookbook \texttt{mobile\_use} tool. The MCP server exposes typed, app-specific operations such as \texttt{caltrack.log\_food} and \texttt{mybank.send\_zelle} over the same 26 apps. We release the MCP server alongside the benchmark so others can run the same comparison.

\begin{table}[t]
\centering
\small
\begin{tabular}{lcc}
\toprule
Qwen3.5 35B-A3B & Pass rate & Rubric score \\
\midrule
Cookbook \texttt{mobile\_use} (7 actions) & 12.8\% & 0.33 \\
Structured per-app tools (MCP)            & 24.8\% & 0.683 \\
\midrule
$\Delta$ & +12.0 pp & +0.353 \\
\bottomrule
\end{tabular}
\caption{\small Qwen3.5 with and without MCP tools over the same 133-task suite.}
\label{tab:mcp-ablation}
\end{table}

\paragraph{Result.} Structured tools raise Qwen3.5's pass rate from 12.8\% to 24.8\% and its mean rubric score from 0.33 to 0.683, but the baseline still trails the frontier models.

\begin{figure*}[t]
\centering
\includegraphics[width=\textwidth,keepaspectratio]{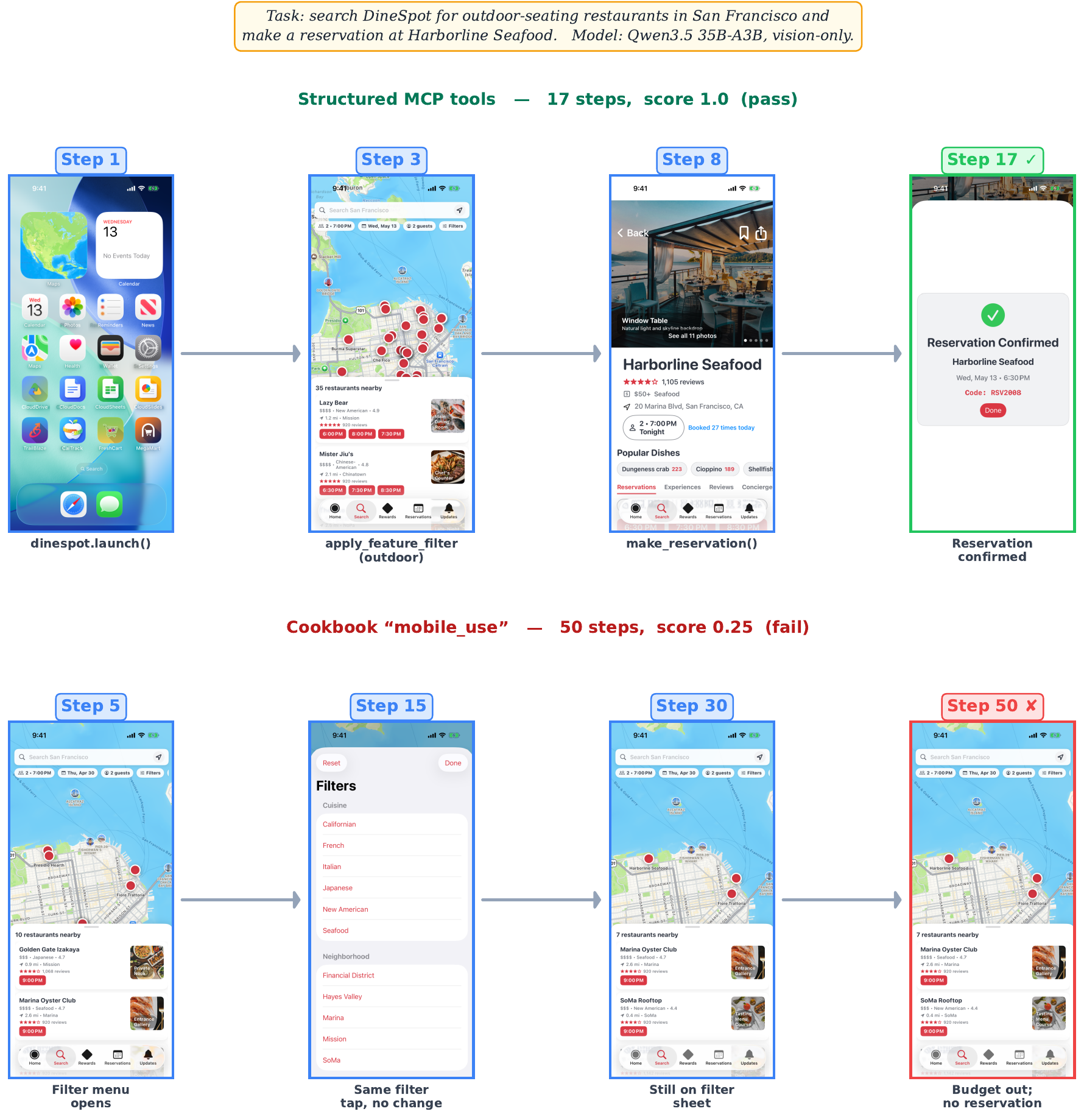}
\caption{\small \texttt{dinespot-001}, Qwen3.5 vision-only. \textbf{Top, structured MCP tools (17 steps, score 1.0).} Typed calls take the agent directly to a confirmed Harborline Seafood booking. \textbf{Bottom, cookbook \texttt{mobile\_use} (50 steps, score 0.25).} The same model gets stuck on the filter sheet and never makes a reservation.}
\label{fig:mcp-traj}
\end{figure*}

%% file: colm2026_conference.bib
@article{kaelbling1998pomdp,
  title={Planning and Acting in Partially Observable Stochastic Domains},
  author={Kaelbling, Leslie Pack and Littman, Michael L. and Cassandra, Anthony R.},
  journal={Artificial Intelligence},
  volume={101},
  pages={99--134},
  year={1998}
}

@article{shi2017world,
  title={World of Bits: An Open-Domain Platform for Web-Based Agents},
  author={Shi, Tianlin and Karpathy, Andrej and Fan, Linxi and Hernandez, Jonathan and Liang, Percy},
  journal={ICML},
  year={2017}
}

@article{liu2018reinforcement,
  title={Reinforcement Learning on Web Interfaces Using Workflow-Guided Exploration},
  author={Evan Zheran Liu and Kelvin Guu and Panupong Pasupat and Tianlin Shi and Percy Liang},
  journal={ICLR},
  year={2018}
}

@article{yao2022webshop,
  title={WebShop: Towards Scalable Real-World Web Interaction with Grounded Language Agents},
  author={Shunyu Yao and Howard Chen and John Yang and Karthik Narasimhan},
  journal={NeurIPS},
  year={2022}
}

@article{deng2023mind2web,
  title={Mind2Web: Towards a Generalist Agent for the Web},
  author={Xiang Deng and Yu Gu and Boyuan Zheng and Shijie Chen and Samuel Stevens and Boshi Wang and Huan Sun and Yu Su},
  journal={NeurIPS},
  year={2023}
}

@article{zhou2023webarena,
  title={WebArena: A Realistic Web Environment for Building Autonomous Agents},
  author={Shuyan Zhou and Frank F. Xu and Hao Zhu and Xuhui Zhou and Robert Lo and Abishek Sridhar and Xianyi Cheng and Tianyue Ou and Yonatan Bisk and Daniel Fried and Uri Alon and Graham Neubig},
  journal={ICLR},
  year={2024}
}

@article{koh2024visualwebarena,
  title={VisualWebArena: Evaluating Multimodal Agents on Realistic Visual Web Tasks},
  author={Jing Yu Koh and Robert Lo and Lawrence Jang and Vikram Duvvur and Ming Chong Lim and Po-Yu Huang and Graham Neubig and Shuyan Zhou and Ruslan Salakhutdinov and Daniel Fried},
  journal={ACL},
  year={2024}
}

@article{he2024webvoyager,
  title={WebVoyager: Building an End-to-End Web Agent with Large Multimodal Models},
  author={Hongliang He and Wenlin Yao and Kaixin Ma and Wenhao Yu and Yong Dai and Hongming Zhang and Zhenzhong Lan and Dong Yu},
  journal={ACL},
  year={2024}
}

@article{xue2025illusion,
  title={An Illusion of Progress? Assessing the Current State of Web Agents},
  author={Tianci Xue and Weijian Qi and Tianneng Shi and Chan Hee Song and Boyu Gou and Dawn Song and Huan Sun and Yu Su},
  journal={COLM},
  year={2025}
}

@misc{kong2025mobileworldbenchmarkingautonomousmobile,
      title={MobileWorld: Benchmarking Autonomous Mobile Agents in Agent-User Interactive and MCP-Augmented Environments}, 
      author={Quyu Kong and Xu Zhang and Zhenyu Yang and Nolan Gao and Chen Liu and Panrong Tong and Chenglin Cai and Hanzhang Zhou and Jianan Zhang and Liangyu Chen and Zhidan Liu and Steven Hoi and Yue Wang},
      year={2025},
      journal={arXiv preprint arXiv:2512.19432}
}

@article{xie2024osworld,
  title={OSWorld: Benchmarking Multimodal Agents for Open-Ended Tasks in Real Computer Environments},
  author={Tianbao Xie and Danyang Zhang and Jixuan Chen and Xiaochuan Li and Siheng Zhao and Ruisheng Cao and Toh Jing Hua and Zhoujun Cheng and Dongchan Shin and Fangyu Lei and Yitao Liu and Yiheng Xu and Shuyan Zhou and Silvio Savarese and Caiming Xiong and Victor Zhong and Tao Yu},
  journal={NeurIPS},
  year={2024}
}

@article{bonatti2024windows,
  title={Windows Agent Arena: Evaluating Multi-Modal OS Agents at Scale},
  author={Rogerio Bonatti and Dan Zhao and Francesco Bonacci and Dillon Dupont and Sara Abdali and Yinheng Li and Yadong Lu and Justin Wagle and Kazuhito Koishida and Arthur Bucker and Lawrence Jang and Zack Hui},
  journal={ICML},
  year={2025}
}

@article{yang2025macosworld,
  title={macOSWorld: A Multilingual Interactive Benchmark for GUI Agents},
  author={Pei Yang and Hai Ci and Mike Zheng Shou},
  journal={NeurIPS},
  year={2025}
}

@article{drouin2024workarena,
  title={WorkArena: How Capable Are Web Agents at Solving Common Knowledge Work Tasks?},
  author={Alexandre Drouin and Maxime Gasse and Massimo Caccia and Issam H. Laradji and Manuel Del Verme and Tom Marty and Léo Boisvert and Megh Thakkar and Quentin Cappart and David Vazquez and Nicolas Chapados and Alexandre Lacoste},
  journal={ICML},
  year={2024}
}

@article{mialon2023gaia,
  title={GAIA: A benchmark for General AI Assistants},
  author={Grégoire Mialon and Clémentine Fourrier and Craig Swift and Thomas Wolf and Yann LeCun and Thomas Scialom},
  journal={ICLR},
  year={2024}
}

@article{xu2025theagentcompany,
  title={TheAgentCompany: Benchmarking LLM Agents on Consequential Real World Tasks},
  author={Frank F. Xu and Yufan Song and Boxuan Li and Yuxuan Tang and Kritanjali Jain and Mengxue Bao and Zora Z. Wang and Xuhui Zhou and Zhitong Guo and Murong Cao and Mingyang Yang and Hao Yang Lu and Amaad Martin and Zhe Su and Leander Maben and Raj Mehta and Wayne Chi and Lawrence Jang and Yiqing Xie and Shuyan Zhou and Graham Neubig},
  journal={NeurIPS},
  year={2025}
}

@article{yao2024taubench,
  title={$\tau$-bench: A Benchmark for Tool-Agent-User Interaction in Real-World Domains},
  author={Shunyu Yao and Noah Shinn and Pedram Razavi and Karthik Narasimhan},
  journal={ICLR},
  year={2025}
}

@article{toyama2021androidenv,
  title={AndroidEnv: A Reinforcement Learning Platform for Android},
  author={Daniel Toyama and Philippe Hamel and Anita Gergely and Gheorghe Comanici and Amelia Glaese and Zafarali Ahmed and Tyler Jackson and Shibl Mourad and Doina Precup},
  journal={arXiv preprint arXiv:2105.13231},
  year={2021}
}

@article{rawles2024aitw,
  title={Android in the Wild: A Large-Scale Dataset for Android Device Control},
  author={Christopher Rawles and Alice Li and Daniel Rodriguez and Oriana Riva and Timothy Lillicrap},
  journal={NeurIPS},
  year={2023}
}

@article{rawles2025androidworld,
  title={AndroidWorld: A Dynamic Benchmarking Environment for Autonomous Agents},
  author={Christopher Rawles and Sarah Clinckemaillie and Yifan Chang and Jonathan Waltz and Gabrielle Lau and Marybeth Fair and Alice Li and William Bishop and Wei Li and Folawiyo Campbell-Ajala and Daniel Toyama and Robert Berry and Divya Tyamagundlu and Timothy Lillicrap and Oriana Riva},
  journal={ICLR},
  year={2025}
}

@article{yang2023appagent,
  title={AppAgent: Multimodal Agents as Smartphone Users},
  author={Chi Zhang and Zhao Yang and Jiaxuan Liu and Yucheng Han and Xin Chen and Zebiao Huang and Bin Fu and Gang Yu},
  journal={CHI},
  year={2025}
}

@article{hong2023cogagent,
  title={CogAgent: A Visual Language Model for GUI Agents},
  author={Wenyi Hong and Weihan Wang and Qingsong Lv and Jiazheng Xu and Wenmeng Yu and Junhui Ji and Yan Wang and Zihan Wang and Yuxuan Zhang and Juanzi Li and Bin Xu and Yuxiao Dong and Ming Ding and Jie Tang},
  journal={CVPR},
  year={2024}
}

@article{wang2024mobileagent,
  title={Mobile-Agent: Autonomous Multi-Modal Mobile Device Agent with Visual Perception},
  author={Junyang Wang and Haiyang Xu and Jiabo Ye and Ming Yan and Weizhou Shen and Ji Zhang and Fei Huang and Jitao Sang},
  journal={ICLR},
  year={2024}
}

@article{xu2024androidlab,
  title={AndroidLab: Training and Systematic Benchmarking of Android Autonomous Agents},
  author={Yifan Xu and Xiao Liu and Xueqiao Sun and Siyi Cheng and Hao Yu and Hanyu Lai and Shudan Zhang and Dan Zhang and Jie Tang and Yuxiao Dong},
  journal={ACL},
  year={2025}
}

@article{chen2025spabench,
  title={SPA-Bench: A Comprehensive Benchmark for SmartPhone Agent Evaluation},
  author={Jingxuan Chen and Derek Yuen and Bin Xie and Yuhao Yang and Gongwei Chen and Zhihao Wu and Li Yixing and Xurui Zhou and Weiwen Liu and Shuai Wang and Kaiwen Zhou and Rui Shao and Liqiang Nie and Yasheng Wang and Jianye Hao and Jun Wang and Kun Shao},
  journal={ICLR},
  year={2025}
}

@article{lee2025bmoca,
  title={B-MoCA: Benchmarking Mobile Device Control Agents across Diverse Configurations},
  author={Juyong Lee and Taywon Min and Minyong An and Dongyoon Hahm and Haeone Lee and Changyeon Kim and Kimin Lee},
  journal={CoLLAs},
  year={2025}
}

@article{qin2025uitars,
  title={UI-TARS: Pioneering Automated GUI Interaction with Native Agents},
  author={Yujia Qin and Yining Ye and Junjie Fang and Haoming Wang and Shihao Liang and Shizuo Tian and Junda Zhang and Jiahao Li and Yunxin Li and Shijue Huang and Wanjun Zhong and Kuanye Li and Jiale Yang and Yu Miao and Woyu Lin and Longxiang Liu and Xu Jiang and Qianli Ma and Jingyu Li and Xiaojun Xiao and Kai Cai and Chuang Li and Yaowei Zheng and Chaolin Jin and Chen Li and Xiao Zhou and Minchao Wang and Haoli Chen and Zhaojian Li and Haihua Yang and Haifeng Liu and Feng Lin and Tao Peng and Xin Liu and Guang Shi},
  journal={arXiv preprint arXiv:2501.12326},
  year={2025}
}

@article{wen2024autodroid,
  title={AutoDroid: LLM-powered Task Automation in Android},
  author={Hao Wen and Yuanchun Li and Guohong Liu and Shanhui Zhao and Tao Yu and Toby Jia-Jun Li and Shiqi Jiang and Yunhao Liu and Yaqin Zhang and Yunxin Liu},
  journal={MobiCom},
  year={2024}
}

@article{bai2024digirl,
  title={DigiRL: Training In-The-Wild Device-Control Agents with Autonomous Reinforcement Learning},
  author={Hao Bai and Yifei Zhou and Mert Cemri and Jiayi Pan and Alane Suhr and Sergey Levine and Aviral Kumar},
  journal={NeurIPS},
  year={2024}
}

@article{bai2025digiq,
  title={Digi-Q: Learning VLM Q-Value Functions for Training Device-Control Agents},
  author={Hao Bai and Yifei Zhou and Li Erran Li and Sergey Levine and Aviral Kumar},
  journal={ICLR},
  year={2025}
}

@article{lu2025guiodyssey,
  title={GUIOdyssey: A Comprehensive Dataset for Cross-App GUI Navigation on Mobile Devices},
  author={Quanfeng Lu and Wenqi Shao and Zitao Liu and Lingxiao Du and Fanqing Meng and Boxuan Li and Botong Chen and Siyuan Huang and Kaipeng Zhang and Ping Luo},
  journal={ICCV},
  year={2025}
}

@article{you2024ferretui,
  title={Ferret-UI: Grounded Mobile UI Understanding with Multimodal LLMs},
  author={Keen You and Haotian Zhang and Eldon Schoop and Floris Weers and Amanda Swearngin and Jeffrey Nichols and Yinfei Yang and Zhe Gan},
  journal={ECCV},
  year={2024}
}

@article{zheng2023judgingllmasajudgemtbenchchatbot,
      title={Judging LLM-as-a-Judge with MT-Bench and Chatbot Arena}, 
      author={Lianmin Zheng and Wei-Lin Chiang and Ying Sheng and Siyuan Zhuang and Zhanghao Wu and Yonghao Zhuang and Zi Lin and Zhuohan Li and Dacheng Li and Eric P. Xing and Hao Zhang and Joseph E. Gonzalez and Ion Stoica},
      year={2023},
      journal={NeurIPS}

}

@misc{anthropic2026opus46,
  title={Claude Opus 4.6},
  author={{Anthropic}},
  year={2026},
  howpublished={\url{https://www.anthropic.com/news/claude-opus-4-6}}
}

@misc{anthropic2026ClaudeCode,
  title={Claude Code},
  author={{Anthropic}},
  year={2026},
  howpublished={\url{https://docs.anthropic.com/en/docs/claude-code/overview}}
}

@misc{openai2026gpt54,
  title={GPT-5.4},
  author={{OpenAI}},
  year={2026},
  howpublished={\url{https://developers.openai.com/api/docs/models/gpt-5.4}}
}

@misc{google2026gemini,
  title={Gemini 3 Flash},
  author={{Google}},
  year={2026},
  howpublished={\url{https://blog.google/products/gemini/gemini-3-flash/}}
}

@misc{qwen2026qwen35,
  title={Qwen3.5-35B-A3B},
  author={{Qwen Team}},
  year={2026},
  howpublished={\url{https://huggingface.co/Qwen/Qwen3.5-35B-A3B}}
}

@article{shen2019skillbot,
  title={SkillBot: Towards Automatic Skill Development via User Demonstration},
  author={Yilin Shen and Avik Ray and Hongxia Jin and Sandeep Nama},
  journal={ACL},
  year={2019}
}
